\RequirePackage{fix-cm}

\documentclass[twocolumn]{svjour3}          
\smartqed  
\usepackage{mathtools}
\usepackage{times}
\usepackage{epsfig}
\usepackage{graphicx}
\usepackage{amssymb}
\usepackage{multirow}
\usepackage{algorithm}
\usepackage[]{algpseudocode}
\usepackage{dsfont}
\usepackage{url}
\usepackage{natbib}
\usepackage{color}
\journalname{International Journal of Computer Vision}
\begin{document}
	
\title{Deep Appearance Models: A Deep Boltzmann Machine Approach for Face Modeling
}
	
\author{Chi Nhan Duong         \and
		Khoa Luu \and
		Kha Gia Quach \and
		Tien D. Bui
	}
	
\institute{
		Chi Nhan Duong $^{1,2}$, Khoa Luu $^{2}$, Kha Gia Quach $^{1,2}$, Tien D. Bui $^{1}$\\
		$^{1}$ Concordia University, Computer Science and Software Engineering, Montr\'eal, Qu\'ebec, Canada.\\
		$^{2}$  CyLab Biometrics Center and the Department of Electrical and Computer Engineering, Carnegie Mellon University, Pittsburgh, PA, USA. \\
		\email{\{c\_duon, k\_q, bui\}@encs.concordia.ca;kluu@andrew.cmu.edu}\\
	}
		\date{Received: date / Accepted: date}

\maketitle

\begin{abstract}
The ``interpretation through synthesis'' approach to analyze face images, particularly Active Appearance Models (AAMs) method, has become one of the most successful face modeling approaches over the last two decades. AAM models have ability to represent face images through synthesis using a controllable parameterized Principal Component Analysis (PCA) model. However, the accuracy and robustness of the synthesized faces of AAM are highly depended on the training sets and inherently on the generalizability of PCA subspaces. This paper presents a novel Deep Appearance Models (DAMs) approach, an efficient replacement for AAMs, to accurately capture both shape and texture of face images under large variations. In this approach, three crucial components represented in hierarchical layers are modeled using the Deep Boltzmann Machines (DBM) to robustly capture the variations of facial shapes and appearances. DAMs are therefore superior to AAMs in inferencing a representation for new face images under various challenging conditions. The proposed approach is evaluated in various applications to demonstrate its robustness and capabilities, i.e. facial super-resolution reconstruction, facial off-angle reconstruction or face frontalization, facial occlusion removal and age estimation using challenging face databases, i.e. Labeled Face Parts in the Wild (LFPW), Helen and FG-NET.
Comparing to AAMs and other deep learning based approaches, the proposed DAMs achieve competitive results in those applications, thus this showed their advantages in handling occlusions, facial representation, and reconstruction.

\end{abstract}

\begin{figure}[t]
	\begin{center}
		\includegraphics[width=8.3cm]{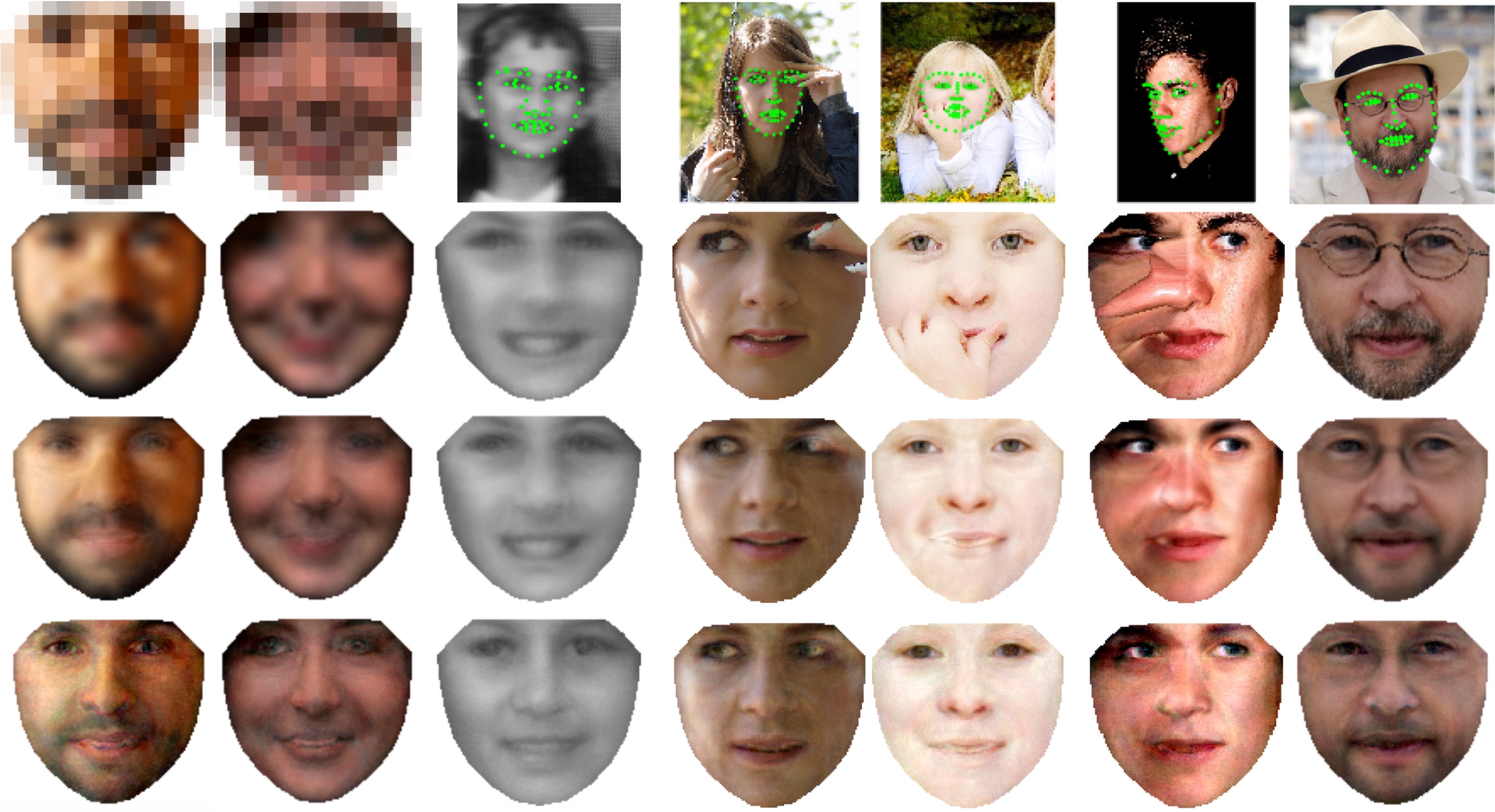}\end{center}
	\caption{An illustration in facial interpretation using the AAMs and our DAMs approach in real world images, e.g. low resolution, blurred faces, occlusions, off-angle faces, etc. The first row: original images; The second row: shape free images; The third row: facial interpretation using PCA-based AAMs; The fourth row: facial interpretation using our proposed DAMs approach.}
	\label{fig:ExtremeCaseDAM_known}
	\label{fig:onecol}
\end{figure}

\section{Introduction}
\label{sec:introduction}
{M}{odeling}  faces with large variations has been a challenging task in computer vision. These variations such as expressions, poses and occlusions are usually complex and non-linear. Moreover, new facial images also come with their own characteristic artifacts greatly diverse. Therefore, a good face modeling approach needs to be carefully designed for flexibly adapting to these challenging issues.
Over the last two decades, the ``interpretation through synthesis'' approach has become one of the most successful and popular face modeling approaches.
This approach aims to ``describe'' a given face image by generating a new synthesized image similar to it as much as possible. This purpose can be achieved by an optimization process on the appearance parameters of the model based \textit{apriori} on constrained solutions. The subspace model then plays a key role that decides the robustness of the whole system. Therefore, in order to be applicable, it must provide a basis for a broad range of variations that are usually unseen.

Active Appearance Models (AAMs), one of the most successful face interpretation methods, were first introduced by 
\cite{CootesAAM1}
. Since then, it has been widely applied in many applications such as face recognition ~\citep{edwards1998face}, facial expression recognition ~\citep{sung2008pose}, face tracking ~\citep{zhu2006real},  
expressive visual text-to-speech ~\citep{anderson2013expressive} and many other tasks. Although the framework of AAMs is general and effective, their generalization ability is still limited especially when dealing with unseen variations.
\cite{gross2005generic} showed that AAMs perform well in person-specific cases rather than generic ones. 
\cite{cootes2006algorithm} pointed out the problem of the pre-computed Jacobian matrix computed during the training step. Since it is only an approximation for testing image, it may lead to poor convergence when the image is very different from training data. Lighting changes ~\citep{pizarro2008light} also make AAMs difficult to synthesize new images.

To overcome these disadvantages, there have been numerous improvements and adaptations based on the original approach ~\citep{matthews2004active, donner2006fast, amberg2009compositional, medina2014BayesianAAM}.
However, even when these adaptations are taken into account, the capabilities of facial generalization and reconstruction are still highly dependent on the characteristics of training databases. This is because at the heart of AAMs, Principal Component Analysis (PCA) is used to provide a subspace to model variations in training data. The limitation of PCA to generalize to illumination and poses, particularly for faces, is very well known. Therefore, it is not surprising that AAMs have difficulties in generalizing to new faces under these challenging conditions.
On the other hand, the variations in data are not only large but also non-linear. For example, the variations in different facial expressions or poses are non-linear. It apparently violates the linear assumptions of PCA-based models. Thus, single PCA model is unable to interpret the facial variations well. Figure \ref{fig:ExtremeCaseDAM_known} presents some faces with various challenging factors, i.e. low-resolution, blurred faces, occlusions, poses. The AAM interpretations presented in the third row of the figure have a major negative impact from these wide range of variations.

Recently, Deep Boltzmann Machines (DBM) ~\citep{salakhutdinov2009deep} have gained significant attention as one of the emerging  research topics in both the higher-level representation of data and the distribution of observations. In DBM, non-linear latent variables are organized in multiple connected layers in a way that variables in one layer can simultaneously contribute to the probabilities or states of variables in the next layers. Each layer learns a different factor to represent the variations in a given data. Thanks to the nonlinear structure of DBM and the strength of latent variables organized in hidden layers, it efficiently captures variations and structures in complex data that could be higher than second order.

Moreover, DBM is shown to be more robust with ambiguous input data \citep{salakhutdinov2009deep}.
There are some recent works using DBM as shape prior model \citep{eslami2014shape,wu2013facial,taylor2010dynamical}.
Far apart from these methods, the higher-level relationships of both shape and texture are exploited in our proposed DAMs so that the reconstruction of one can benefit from the information on the other.
This paper proposes a novel Deep Appearance Models (DAMs\footnote{Noted that the term DAM is also used for "Direct Appearance Models" in ~\citep{hou2001direct}.}) approach
to find a set of parameters in both shape and texture to characterize the identity, facial poses, facial expressions, lighting conditions of a given face.
In addition, our proposed approach also has ability to generate a compact set of parameters in a robust model that can later be used for classification.
Specifically, the DBM-based shape and texture models are first independently constructed.
Then the interactions between these shapes and textures are further modeled using a deeper hidden layer. By this way, after fitting the model to new images, these interactions can be used as a compact set of parameters that represent both shape and appearance of faces for further discriminative problems.
Furthermore, unlike other deep learning based approaches such as CNNs, with a specific topology of a stochastic neural network and sampling based weight update process (i.e. Contrastive Divergence), the need for large-scale training data is also alleviated in our DAMs structure.

A preliminary version of our work can be found in \citep{Duong_2015_CVPR}. In that work, our proposed DAMs \footnote{The implementation of DAMs will be available at \url{https://github.com/dcnhan/DeepAppearanceModels} and our project page \url{http://www.contrib.andrew.cmu.edu/~kluu/faceaging.html} } are able to capture a wide range of face variations as well as efficiently interpret the connections between face shape and texture. In this paper, we further extend the fitting stage to make it more robust to occlusions and noise. As a result, the model represented in this work is more advanced in terms of both face representation and model fitting.
The paper is organized as follows. Section \ref{sec:RelatedWork} reviews some recent AAMs-based and deep learning based approaches in both face representation and modeling. Section \ref{sec:DAMs} presents the structure of DAMs and their properties. The model fitting is given in Section \ref{sec:Model_fitting} followed by the experimental results in Section \ref{sec:experiment}. Finally, the conclusion together with future work are given in Section \ref{sec:Conclusion}.

\section{Related Work} \label{sec:RelatedWork}

This section briefly reviews recent advances of AAMs-based approaches for constructing and fitting deformable models, and deep learning based methods for modeling human faces.

\subsection{AAMs Modeling and Fitting}
The basic AAMs \citep{cootes2001active} 
build a statistical unified appearance model describing both shape and texture variation.
One of the major drawbacks of AAMs is that the models only capture small amounts of appearance variations which can lead to poor performance on unknown variations caused by changes in the real-world environment, e.g. poses, lighting and camera conditions. 
~The second drawback is that \emph{person specific} AAMs substantially outperform generic AAMs trained across numerous subjects.
~Addressing the first drawback, some improvements have been made by applying the ideas of mixture models \citep{van2010capturing} and probabilistic PCAs \citep{medina2014BayesianAAM} to represent as much variations as possible especially in the appearance model.
Descriptive feature-based approaches were employed instead of intensity-based to deal with the second drawback of AAMs.
~\cite{ge2013active} proposed three Gabor-based texture representations for AAMs capturing the properties of both Gabor magnitude and phase. 
These Gabor-based texture representations are more compact 
and robust to various conditions, e.g. expression, illumination and pose changes. 
\cite{antonakos2014hog} proposed to use dense histogram of oriented gradients features with AAMs to enhance their robustness and performance on unseen faces. 
\cite{haase2014instance} proposed a transfer learning based approach which incorporates related knowledge obtained from another training set with unseen illumination conditions to the existing AAMs to improve their generalization ability. 
Fitting steps in AAMs are an iterative optimization process measuring the cost between a testing image and a model texture in the coordinate of a reference frame.
Generally, previous fitting techniques can be divided into two categories, i.e. discriminative and generative approaches.
In the first category, the optimizing process is updated using a train-ed parameter-updating model.
The model can be trained in several ways, e.g. perturbing the parameters and recording the residuals ~\citep{CootesAAM1}, directly using texture information to predict the shape ~\citep{hou2001direct}, linear regression ~\citep{donner2006fast} and non-linear regression methods ~\citep{saragih2007nonlinear}.
These techniques usually require low computational costs but their quality is still limited since the mapping function is fixed and independent of current model parameters.
In the second category, the fitting steps are formulated as an image alignment problem and iteratively solved via the Gaussian-Newton optimization technique.
\cite{matthews2004active} presented a project out inverse algorithm to work on the orthogonal complement of the texture subspace. 
\cite{navarathna2011fourier} 
proposed a computationally efficient fitting algorithm based on a variant of the Lucas-Kanade (LK) algorithm, called Fourier LK or Fourier AAM, to provide invariance to both expression and illumination.
Other methods find the shape and texture increments
either simultaneously ~\citep{gross2005generic} or alternatively ~\citep{papandreou2008adaptive}. 
\cite{amberg2009compositional}
presented the compositional framework. 
\cite{tzimiropoulos2013optimization} presented a fitting algorithm that works effectively in both forward and inverse cases. 
\cite{mollahosseini2013bidirectional} proposed bidirectional warping method based on image alignment for AAMs fitting.

\subsection{Deep Learning based Approaches}

There has been significant recent interest in deep learning, e.g. deep convolutional networks and stacked auto-encoders, for face modeling or face representation, and face alignment.
\cite{zhu2013deep} proposed to learn the face identity-preserving (FIP) features to represent faces while reducing significantly pose and illumination variances and preserving discriminative features for face recognition task. The authors designed a deep network that contains feature extraction layers, which produce FIP features, and a reconstruction layer, which reconstructs the canonical faces, i.e. frontal faces with neutral expression and normal illumination.  
Later, 
\cite{zhu2014multi} proposed a deep neural network so called multi-view perceptron (MVP) designed to separate the identity and view features. MVP can also generate multi-view images under unobserved viewpoints from a single 2D face image.
\cite{huang2012learning} constructed local convolutional restricted Boltzmann machines that could exploit the global structure while achieving scalability and robustness to small misalignments.
\cite{sun2014deep} proposed to learn a set of high-level features so called Deep hidden IDentity features 
for face verification. 
They trained 60 deep convolutional neural networks (CNNs) to extract complementary and over-complete representations from the last hidden layers of the networks. 
\cite{taigman2014deepface} aim at improving the alignment and the representation step in face verification pipeline by using 3D model-based alignment and a nine-layer CNN. 
Both unsupervised similarity, i.e. the inner vector product,  
and supervised metric, i.e. the $\chi^2$ similarity and the Siamese network are employed.

\cite{kan2014stacked} proposed a deep network called stacked progressive auto-encoders to learn pose-robust features for face recognition by modeling the complex non-linear transformation from the non-frontal face images to frontal ones. 
Similarly, 
\cite{gao2015single} proposed to learn robust face representation features using a deep architecture based on supervised auto-encoder. 
The learning aims at transforming the faces with variants to the canonical view, and extracting similar features from the same subject. 
\cite{ding2015robust} proposed a deep learning framework to jointly learn face representation using multimodal information. The model consists of a set of CNNs to extract complementary facial features, and a three-layer stacked auto-encoder to compress the extracted features. 
Besides learning identity features for face recognition or verification tasks, age-related features could also be extracted ~\citep{zhai2015face, liu2016deep}.
Locating facial key points is also an essential step to represent facial shape.  
\cite{sun2013deep} proposed three-level cascaded CNNs for coarse-to-fine facial point detection (only detecting left eye, right eye, nose, and two mouth corners).

In addition, dealing with noise and occlusions,  
Robust Boltzmann machines (RoBMs) \citep{tang2012robust} were proposed to extend RBM's ability of estimating noise and distinguishing corrupted and uncorrupted pixels to find the optimal latent representations. The structure of RoBM consists of three components: a Gaussian RBM to model the ``clean'' data, a binary RBM for noise modeling, and a multiplicative gating mechanism to separate the clean data from noise/occlusion. Similar structure was used in \citep{tang2012deep} to combine RBM with Lambertian reflectance model including the albedo and surface normals modeling instead of the occlusion/noise modeling.
\cite{li2014intrinsic} presented a single-face image decomposition method 
for image editing operations like relighting and re-texturing. 
They improved decomposition for faces by using human face priors including skin reflectance model and facial geometry. 
\cite{yildirim2015efficient} proposed to combine a generative model based on 3D computer graphics and a discriminative model based on a CNN. This model can reconstruct the approximate shape and texture of a novel face from a single view.

\begin{figure}[t]
	\begin{center}
		\includegraphics[width=8.5cm]{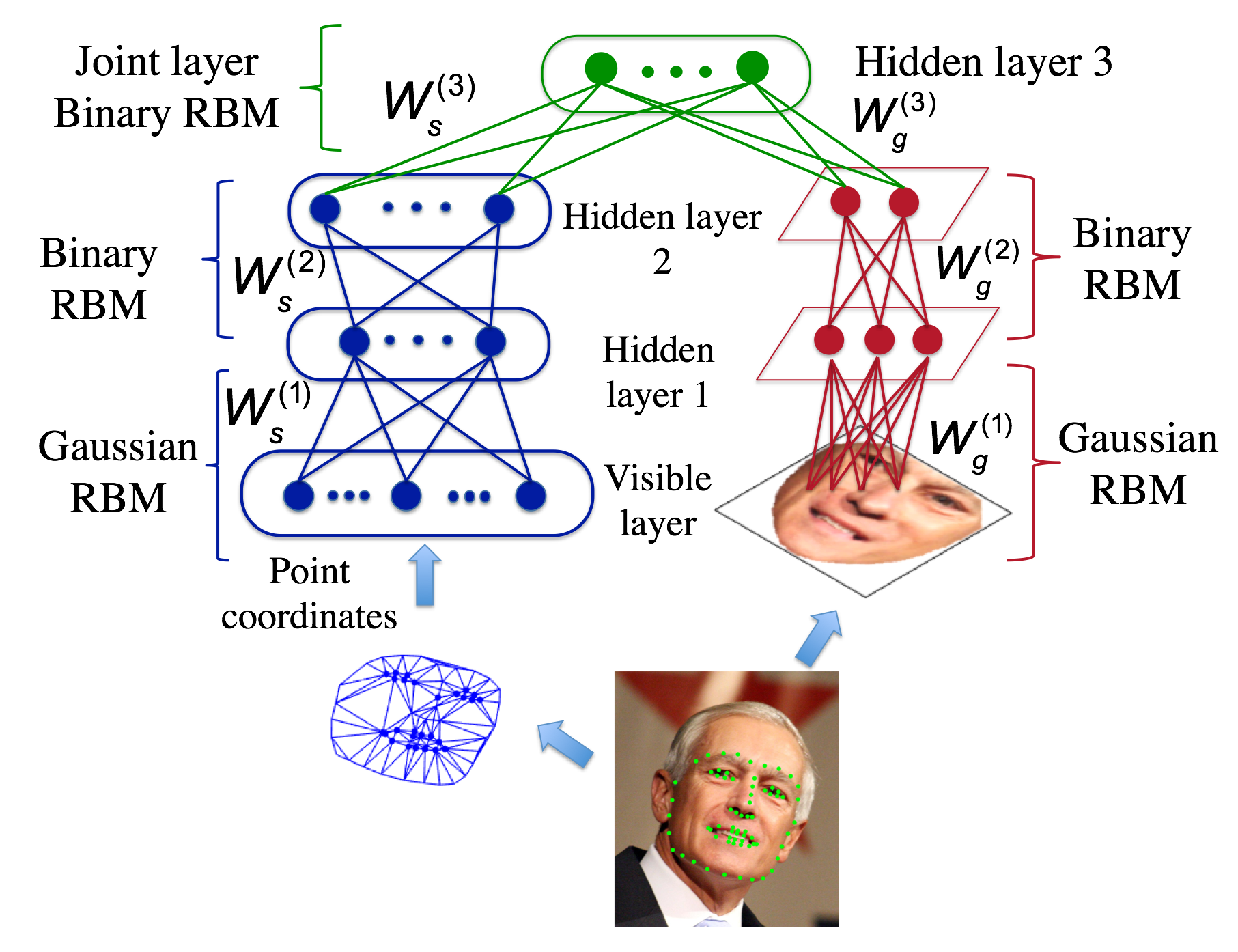}\end{center}
	\caption{Deep Appearance Models that consists of shape model (left), texture model (right) and the joint representation of shape and texture.}
	\label{fig:DBM_based_AAM}
\end{figure}
\section{Deep Appearance Models (DAMs)} \label{sec:DAMs}

The structure of DAMs consists of three main parts, i.e. two prior models for shape and texture and an additional higher-level hidden layer for appearance modeling. The shape model is used to learn the facial shape structure while texture model is used for texture variations. Both of them are mathematically modeled using the Deep Boltzmann Machines that are capable to model high-order correlations among input data. Their undirected connections provide both bottom-up and top-down passes to efficiently send updates between the texture model and the shape model.
These modeling shape and texture parameters are then embedded in a higher-level layer that can be learned by clamping both shapes and textures as observations for the model.
In this section, we present three main steps to construct the model, i.e. shape, texture and appearance modeling.
Then in the next section, a fitting algorithm will be presented in order to synthesize any given new face image.

\subsection{Shape Modeling} \label{subsec:shapemodeling}

In order to generalize possible patterns of facial shapes, we employ a two-layer DBM to learn the distributions of their landmark points. As illustrated in Figure \ref{fig:DBM_based_AAM}, the shape model, i.e. left part of DAMs, consists of a set of visible units encoding the coordinates of landmark points and two sets of hidden units that are latent variables. The connections are symmetric and only those connecting units in adjacent layers are employed.

Let a shape $\mathbf{s}=[x_1,y_1,...,x_N,y_N]^T$ with $N$ landmark points $\{x_i,y_i\},x_i\in\mathbb{R},y_i\in\mathbb{R}$ be the visible units; and $\mathbf{h}^{(1)}_s \in \{0,1\}^{F_s^1},\mathbf{h}^{(2)}_s\in\{0,1\}^{F_s^2}$ be the binary variables of the first and second hidden layers respectively. $F_s^1$ and $F_s^2$ stand for the number of units in these hidden layers.
Since $\{x_i,y_i\}$  are real values while $\mathbf{h}_s^{(1)}$ and $\mathbf{h}_s^{(2)}$ are binary, we employ the Gaussian-Bernoulli Restricted Boltzmann Machines (GRBM) for the first layer and binary-binary RBM for the subsequent one.
The energy of the joint configuration $\{\mathbf{s},\mathbf{h}^{(1)}_s,\mathbf{h}^{(2)}_s\}$ in facial shape modeling is formulated as follows:
\small
\begin{equation} \label{eq:ShapeDBM}
\begin{split}
E(\mathbf{s},\mathbf{h}^{(1)}_s,\mathbf{h}^{(2)}_s;\theta_s)=&\sum_{i}{\frac{(s_i-b_{s_i})^2}{2\sigma^2_{s_i}}}-\sum_{i,j}{\frac{s_i}{\sigma_{s_i}}W^{(1)}_{sij}h^{(1)}_{sj}}\\
&-\sum_{j,l}{h^{(1)}_{sj}W^{(2)}_{sjl}h^{(2)}_{sl}}
\end{split}
\end{equation}
\normalsize
where $\theta_s=\{\mathbf{W}^{(1)}_s,\mathbf{W}^{(2)}_s,\boldsymbol{\sigma}_s^2, \mathbf{b}_s \}$ are the model parameters representing connecting weights of visible-to-hidden and hidden-to-hidden interactions, the variance, and the bias of visible units.

Notice that in Eqn. \eqref{eq:ShapeDBM}, the bias terms of hidden units are ignored to simplify the equation. Its corresponding probability is then given by the Boltzmann distribution:
\small
\begin{equation}
\begin{split}
P(\mathbf{s};\theta_s)&=\sum_{\mathbf{h}^{(1)}_s,\mathbf{h}^{(2)}_s}{P(\mathbf{s},\mathbf{h}^{(1)}_s,\mathbf{h}^{(2)}_s;\theta_s)}\\
&=\frac{1}{Z(\theta_s)}\sum_{\mathbf{h}^{(1)}_s,\mathbf{h}^{(2)}_s}{e^{-E(\mathbf{s},\mathbf{h}^{(1)}_s,\mathbf{h}^{(2)}_s;\theta_s)}}
\end{split}
\end{equation}
\normalsize
where $Z(\theta_s)$ is the partition function.

\begin{figure}[!t]
	\begin{center}
		\includegraphics[width=8.6cm]{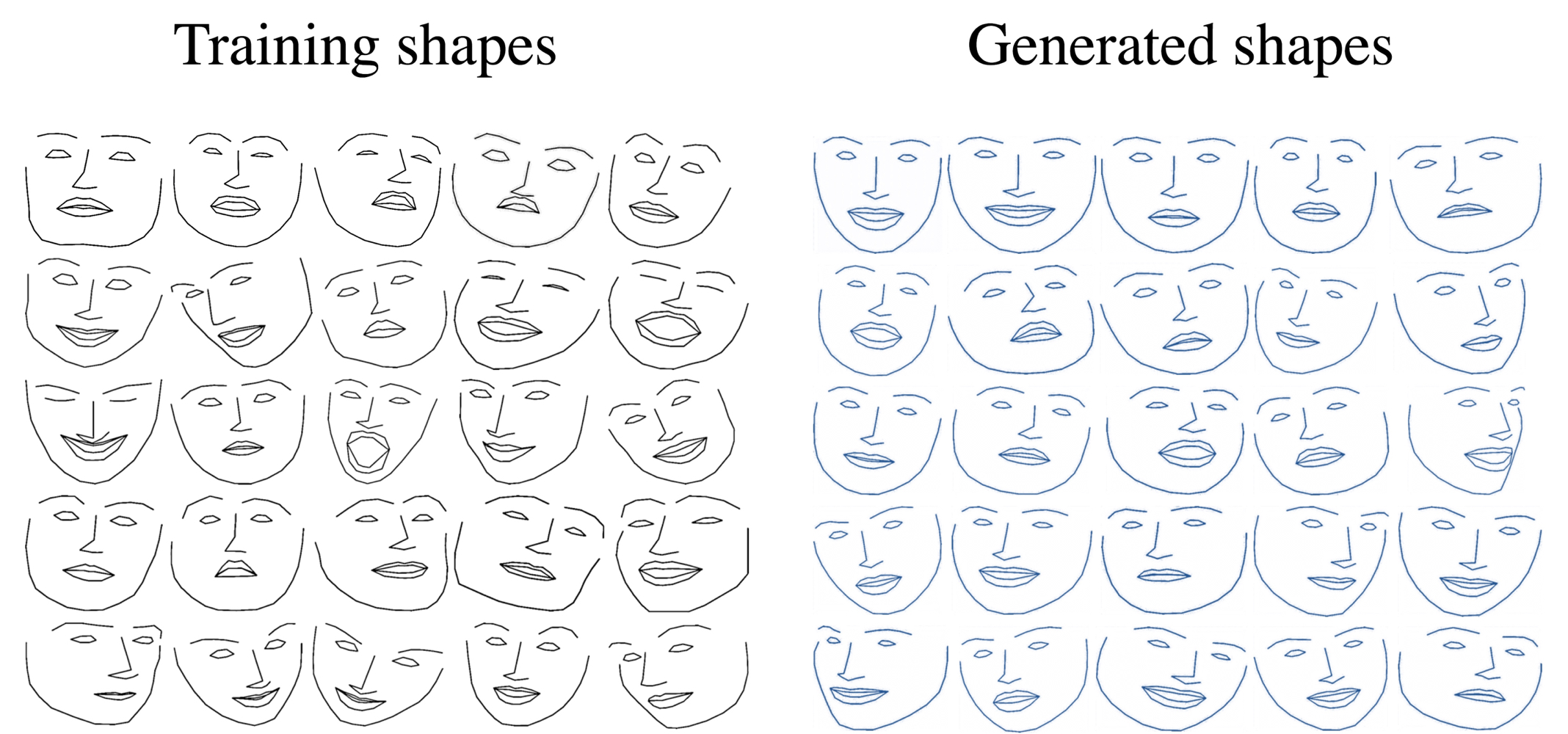}\end{center}
	\caption{A subset of training shapes and generated shapes from shape model with 10-step Gibbs sampling.}
	\label{fig:DAM_generated_shape}
\end{figure}

The conditional distributions over $\mathbf{s}, \mathbf{h}^{(1)}_s$, and $\mathbf{h}^{(2)}_s$ are then given as in Eqn. (\ref{eq:Shape_Cond_Dist}).
\small
\begin{equation} \label{eq:Shape_Cond_Dist}
\begin{split}
p(h_{sj}^{(1)}|\mathbf{s},\mathbf{h}_s^{(2)}) &= \delta\left(\sum_{i}W_{sij}^{(1)}\frac{s_i}{\sigma_{s_i}}+\sum_l W_{sjl}^{(2)}h_{sl}^{(2)}\right)\\
p(h_{sl}^{(2)}|\mathbf{h}_s^{(1)}) &= \delta\left(\sum_j W_{sjl}^{(2)}h_{sj}^{(1)}\right)\\
s_i|\mathbf{h}_s^{(1)} &\sim\mathcal{N}\left(\sigma_{s_i}\sum_j {W_{sij}^{(1)} h_{sj}^{(1)}} + b_{s_i},\sigma_{s_i}^2\right)
\end{split}
\end{equation}
\normalsize
where $\delta(x)=1/(1+\exp(-x))$ is the logistic function.
Figure \ref{fig:DAM_generated_shape} illustrates a subset of training shapes together with  samples generated from shape model after 10-step Gibbs sampling. From this, one can see that the shape model is able to capture the overall shape structure as well as a wide range of head poses and expressions.

\subsection{Texture Modeling} \label{subsec:texmodeling}
As opposed to facial shapes, the appearance of human face usually varies drastically due to numerous factors such as identities, lighting conditions, facial occlusions, expressions, image resolutions, etc. These factors can significantly change pixel values presented in these textures and result in much higher non-linear variations.
Therefore, the process of texture modeling is more complicated and requires the texture model to be sophisticated enough to represent the variations.

The structure of texture model is represented in the right part of DAMs in Figure \ref{fig:DBM_based_AAM}.
Different from the shape model which directly works with landmark coordinates in image domain $\mathcal{I}\subset\mathds{R}^2$,
the given facial image is first warped from $\mathcal{I}$ to texture domain $\mathcal{D}\subset\mathds{R}^2$ using a reference candidate obtained from the training data. Then the obtained shape-free image is vectorized and used as the visible units for texture model.
The purpose of warping step is to remove the effect of shape factors from the texture model and, therefore, making it more robust to shape changes during modeling process.
Specifically, given an image $I$, the texture $\mathbf{g}$ is computed as
\begin{equation}
\mathbf{g} = \mathrm{vec}\left(I(W(r_{\mathcal{D}},\mathbf{s}))\right)
\end{equation}
where $\mathrm{vec}(\cdot)$ is the vectorization operator; $W(r_{\mathcal{D}},\mathbf{s})=r_{\mathcal{I}}$ is the warping operator; $r_\mathcal{I} = \left(x_I, y_I\right)$
and $ r_\mathcal{D} = \left(x_i, y_i\right)$ are the 2D locations in image domain $\mathcal{I}$ and texture domain $\mathcal{D}$, respectively.

\begin{figure}[!t]
	\begin{center}
		\includegraphics[width=8.5cm]{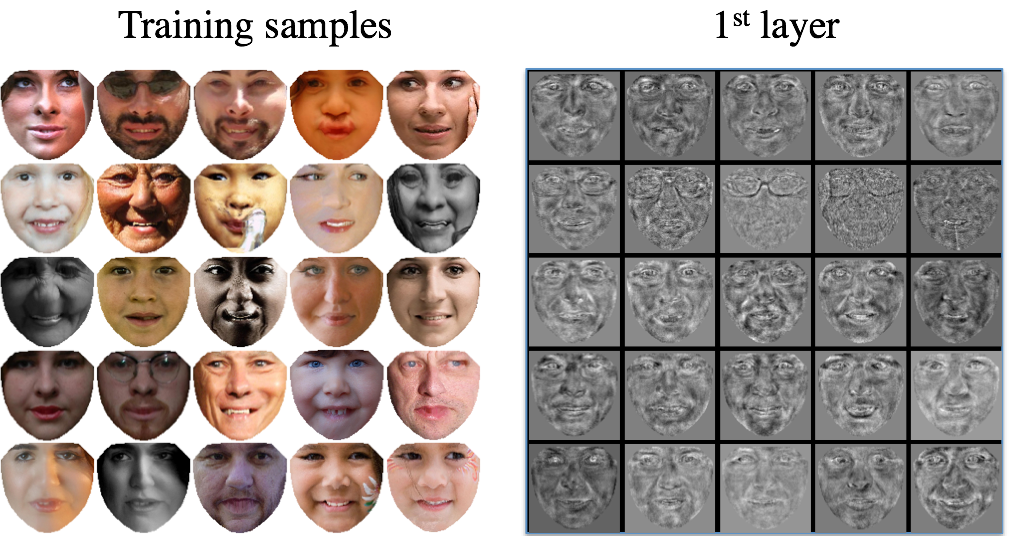}\end{center}
	\caption{A subset of training faces and learned features of the first layer texture model.}
	\label{fig:DAM_learned_feature}
\end{figure}

A two-layer DBM is then employed to model the distributions of texture feature represented in $\mathbf{g}$. Similar to shape model, the GRBM is used in the bottom layer while interactions between hidden units in higher layers are formulated by a binary-binary RBM.
The energy of a state $\{\mathbf{g},\mathbf{h}^{(1)}_g,\mathbf{h}^{(2)}_g\}$ in texture modeling is given as in Eqn. (\ref{eq:TextureDBM}) where $\{\mathbf{h}^{(1)}_g,\mathbf{h}^{(2)}_g\}$ denote the set of hidden units and $\theta_g=\{\mathbf{W}^{(1)}_g,\mathbf{W}^{(2)}_g,\boldsymbol{\sigma}_g^2,\mathbf{b}_g\}$ are the model parameters.

\small
\begin{eqnarray} \label{eq:TextureDBM}
\begin{aligned}
E(\mathbf{g},\mathbf{h}^{(1)}_g,\mathbf{h}^{(2)}_g;\theta_g)=&\sum_{k}{\frac{(g_k-b_{g_k})^2}{2\sigma^2_{g_k}}}-\sum_{k,t}{\frac{g_k}{\sigma_{g_k}}W^{(1)}_{gkt}h^{(1)}_{gt}}\\
&-\sum_{t,v}{h^{(1)}_{gt}W^{(2)}_{gtv}h^{(2)}_{gv}}
\end{aligned}
\end{eqnarray}
\normalsize
The probability of $\mathbf{g}$ assigned by the model is computed as:
\small
\begin{equation}
\begin{split}
P(\mathbf{g};\theta_g)&=\sum_{\mathbf{h}^{(1)}_g,\mathbf{h}^{(2)}_g}{P(\mathbf{g},\mathbf{h}^{(1)}_g,\mathbf{h}^{(2)}_g;\theta_g)}\\
&=\frac{1}{Z(\theta_g)}\sum_{\mathbf{h}^{(1)}_g,\mathbf{h}^{(2)}_g}{e^{-E(\mathbf{g},\mathbf{h}^{(1)}_g,\mathbf{h}^{(2)}_g;\theta_g)}}
\end{split}
\end{equation}
\normalsize

The conditional distributions over $\mathbf{g}$, $\mathbf{h}_g^{(1)}$, and $\mathbf{h}_g^{(2)}$ are derived similar to those of shape model as in Eqn. (\ref{eq:Texture_Cond_Dist}). Figure \ref{fig:DAM_learned_feature} illustrates a subset of training texture and the learned feature obtained using the first layer of the presented texture model.
\small
\begin{equation} \label{eq:Texture_Cond_Dist}
\begin{split}
p(h_{gt}^{(1)}|\mathbf{g},\mathbf{h}_g^{(2)}) &= \delta\left(\sum_{k}W_{gkt}^{(1)}\frac{g_k}{\sigma_{g_k}}+\sum_v W_{gtv}^{(2)}h_{gv}^{(2)}\right)\\
p(h_{gv}^{(2)}|\mathbf{h}_g^{(1)}) &= \delta\left(\sum_t W_{gtv}^{(2)}h_{gt}^{(1)}\right)\\
g_k|\mathbf{h}_g^{(1)} &\sim\mathcal{N}\left(\sigma_{g_k}\sum_t {W_{gkt}^{(1)} h_{gt}^{(1)}} + b_{g_k},\sigma_{g_k}^2\right)
\end{split}
\end{equation}
\normalsize

\subsection{Appearance Modeling} \label{subsec:DAMmodeling}

A straightforward way to extract model parameters for both shape and texture is to employ a weighted concatenation and apply a dimensional reduction method such as PCA. However, this is not an optimal solution since these parameters are presented in different domains, i.e. shape parameters $\boldsymbol{\alpha}_s$ determine the coordinates of landmark points while texture parameters $\boldsymbol{\alpha}_g$ present facial appearance in the texture domain $\mathcal{D}$. Therefore, the gaps between them still exist in the final model parameters although weight values are employed to balance the combined features.

Meanwhile, our Deep Appearance Models also aim to produce a robust facial shape and texture representation.
It, however, can be considered as the problem of data learning from multiple sources. In this problem, the information learned from multiple input channels can complement each other and boost the overall performance of the whole model.
Particularly, captions and tags can be used to improve the classification accuracy ~\citep{huiskes2010new, ngiam2011multimodal, srivastava2012multimodal}.

In order to generate a robust feature in DAMs,
one should notice that the hidden units are powerful in terms of increasing the flexibility of deep model.
Besides the ability of capturing different factors from the observations, the higher layer these hidden units are in, the more independent of the specific correlations of an input source ~\citep{srivastava2012multimodal}. Therefore, we can use them as a source-free representation. From that reason, we construct one more high-level layer to interpret the connections between face shape and its texture. Since $\mathbf{h}^{(2)}_s$ and $\mathbf{h}^{(2)}_g$ are independent of the spaces where the coordinates and appearance are in, the new layer can encode the shape and texture information more naturally and effectively.

Let $\mathbf{h}^{(3)}$ be the connection layer and $\theta=\{\theta_s,\theta_g\}$, the energy of the joint configuration $\{\mathbf{s},\mathbf{g},\mathbf{h}_s^{(1)},\mathbf{h}_s^{(2)},\mathbf{h}_g^{(1)},\mathbf{h}_g^{(2)}$, $\mathbf{h}^{(3)}\}$ in DAMs is defined as the summation of three energy functions of shape model, texture model and the joint layer.
\small
\begin{equation}
\begin{split}
& E(\mathbf{s},\mathbf{g},\mathbf{h}_s,\mathbf{h}_g;\theta)\\
&=\sum_{i}{\frac{(s_i-b_{s_i})^2}{2\sigma^2_{s_i}}}-\sum_{i,j}{\frac{s_i}{\sigma_{s_i}}W^{(1)}_{sij}h^{(1)}_{sj}}-\sum_{j,l}{h^{(1)}_{sj}W^{(2)}_{sjl}h^{(2)}_{sl}}\\
&+\sum_{k}{\frac{(g_k-b_{g_k})^2}{2\sigma^2_{g_k}}}-\sum_{k,t}{\frac{g_k}{\sigma_{g_k}}W^{(1)}_{gkt}h^{(1)}_{gt}}-\sum_{t,v}{h^{(1)}_{gt}W^{(2)}_{gtv}h^{(2)}_{gv}}\\
&-\sum_{l,n} h_{sl}^{(2)}W_{sln}^{(3)}h_n^{(3)} -\sum_{v,n} h_{gv}^{(2)}W_{gvn}^{(3)}h_n^{(3)}
\end{split}
\end{equation}
\normalsize
where $\mathbf{h}_s = \{\mathbf{h}_s^{(1)},\mathbf{h}_s^{(2)}\}$ and $\mathbf{h}_g = \{\mathbf{h}_g^{(1)},\mathbf{h}_g^{(2)}\}$.
The joint distribution over the multimodal input can be written as:
\small
\begin{eqnarray}
\begin{aligned}
P(\mathbf{s},\mathbf{g};\theta)&=\sum_{\mathbf{h}^{(2)}_s,\mathbf{h}^{(2)}_g,\mathbf{h}^{(3)}}P(\mathbf{h}^{(2)}_s,\mathbf{h}^{(2)}_g,\mathbf{h}^{(3)})\\
&\left(\sum_{\mathbf{h}^{(1)}_s}P(\mathbf{s},\mathbf{h}^{(1)}_s,\mathbf{h}^{(2)}_s)\right)
\left(\sum_{\mathbf{h}^{(1)}_g}P(\mathbf{g},\mathbf{h}^{(1)}_g,\mathbf{h}^{(2)}_g)\right)
\end{aligned}
\end{eqnarray}
\normalsize
and the conditional distributions over $\mathbf{h}_s^{(2)}$, $\mathbf{h}_g^{(2)}$, and $\mathbf{h}^{(3)}$ are derived as
\small
\begin{equation} \label{eq:DAM_Cond_Dist}
\begin{split}
p(h_{sl}^{(2)}|\mathbf{h}_s^{(1)},\mathbf{h}^{(3)}) &= \delta\left(\sum_j W_{sjl}^{(2)}h_{sj}^{(1)}+\sum_n W_{sln}^{(3)} h_n^{(3)}\right)\\
p(h_{gv}^{(2)}|\mathbf{h}_g^{(1)},\mathbf{h}^{(3)}) &= \delta\left(\sum_t W_{gtv}^{(2)}h_{gt}^{(1)}+\sum_n W_{gvn}^{(3)} h_n^{(3)}\right)\\
p(h_n^{(3)}|\mathbf{h}_s^{(2)},\mathbf{h}_g^{(2)}) &= \delta\left( \sum_l W_{sln}^{(3)} h_{sl}^{(2)} + \sum_v W_{gvn}^{(3)} h_{gv}^{(2)}\right)
\end{split}
\end{equation}
\normalsize
Other conditional distributions over $\mathbf{s}$, $\mathbf{g}$, $\mathbf{h}_s^{(1)}$ and $\mathbf{h}_g^{(1)}$ are the same as in Eqns. (\ref{eq:Shape_Cond_Dist}) and (\ref{eq:Texture_Cond_Dist}).

\begin{figure*} [!t]
	\begin{center}
		\includegraphics[width=13.5cm]{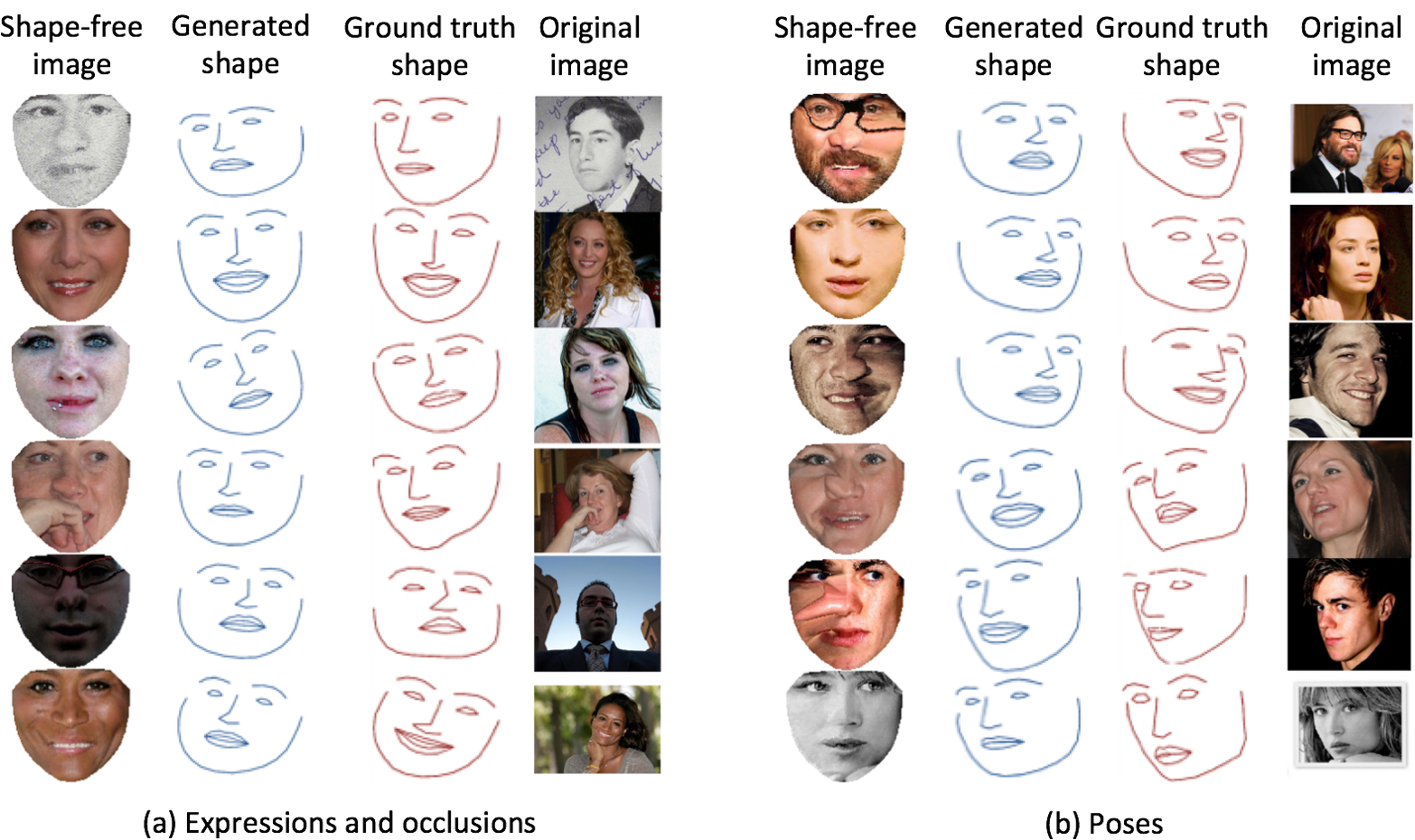}
	\end{center}
	\caption{Facial shape generation using texture information with (a) expressions and occlusions; and (b) poses. In both cases, given the shape-free image (first column), DAMs are able to generate the facial shape (second column) by sampling from $P(\mathbf{s}|\mathbf{g},\theta)$. The ground truth shapes and original images are also given in the third and fourth columns, respectively.}
	\label{fig:generated_shape_given_texture}
\end{figure*}
\subsection{Model Learning}
The parameters in the model are optimized in order to maximize the log likelihood 
$\theta^* = \arg \max_{\theta} \log{P(\mathbf{s},\mathbf{g};\theta)}$.
Then the optimal parameter values can be obtained in a gradient descent fashion given by
\small
\begin{equation} \label{eq:Model_Learning}
\frac{\partial}{\partial \theta}\mathbb{E}\left[\log P(\mathbf{s},\mathbf{g};\theta) \right]=\mathbb{E}_{\text{data}}\left[\frac{\partial E}{\partial \theta}\right]-\mathbb{E}_{\text{model}}\left[\frac{\partial E}{\partial \theta}\right]
\end{equation}
\normalsize
where $\mathbb{E}_{\text{data}}\left[ \cdot \right]$ and $\mathbb{E}_{\text{model}}\left[ \cdot \right]$ are the expectations with respect to data distribution, i.e. \textit{data-dependent expectation}, and distribution estimated by DAM, i.e. \textit{model's expectation}. The former term can be approximated by mean-field inference while the latter term can be estimated using Markov-chain Monte-Carlo (MCMC) based stochastic approximation.

\noindent
\textbf{Computing Data-dependent Expectation:}
Mean-field approximation can be used to compute the first term of Eqn. (\ref{eq:Model_Learning}) ~\citep{salakhutdinov2009deep}.
The main idea of this technique comes from the variational approach where the lower bound of the log-likelihood is maximized with respect to the variational parameters $\boldsymbol{\mu}$.
In the mean-field approximation, for each training face with its shape and texture $\mathbf{s,g}$, all visible units corresponding to $\mathbf{s}$ and $\mathbf{g}$ are fixed and the states of hidden units in the models are set to $\boldsymbol{\mu}$ which are iteratively updated through layers using mean-field fixed-point equations:
\small
\begin{equation} \label{eq:Data_dependent_Expectation}
\begin{split}
\mu_{sj}^{(1)} & \leftarrow \delta\left(\sum_{i}W_{sij}^{(1)}\frac{s_i}{\sigma_{s_i}}+\sum_l W_{sjl}^{(2)}\mu_{sl}^{(2)}\right)\\
\mu_{sl}^{(2)} &\leftarrow \delta\left(\sum_j W_{sjl}^{(2)}\mu_{sj}^{(1)}+\sum_n W_{sln}^{(3)} \mu_n^{(3)}\right)\\
\mu_{gt}^{(1)} &\leftarrow \delta\left(\sum_{k}W_{gkt}^{(1)}\frac{g_k}{\sigma_{g_k}}+\sum_v W_{gtv}^{(2)}\mu_{gv}^{(2)}\right)\\
\mu_{gv}^{(2)} &\leftarrow \delta\left(\sum_t W_{gtv}^{(2)}\mu_{gt}^{(1)}+\sum_n W_{gvn}^{(3)} \mu_n^{(3)}\right)\\
\mu_n^{(3)} &\leftarrow \delta\left( \sum_l W_{sln}^{(3)} \mu_{sl}^{(2)} + \sum_v W_{gvn}^{(3)} \mu_{gv}^{(2)}\right)
\end{split}
\end{equation}
\normalsize
Using these variational parameters, the data-dependent statistics are then computed by averaging over training cases.

\noindent
\textbf{Computing Expectation of the Model:}
For the second term of Eqn. (\ref{eq:Model_Learning}), the MCMC sampling can be applied ~\citep{salakhutdinov2009learning}. Specifically, given the current state of visible and hidden units, their new states are obtained by employing a few steps of persistent Gibbs sampling using Eqns. (\ref{eq:Shape_Cond_Dist}), (\ref{eq:Texture_Cond_Dist}) and (\ref{eq:DAM_Cond_Dist}).
Then the $\mathbb{E}_{\text{model}}\left[ \cdot \right]$ is approximated by the expectations with respect to the new states of the model units.
\begin{figure*} [!t]
	\begin{center}
		\includegraphics[width=16cm]{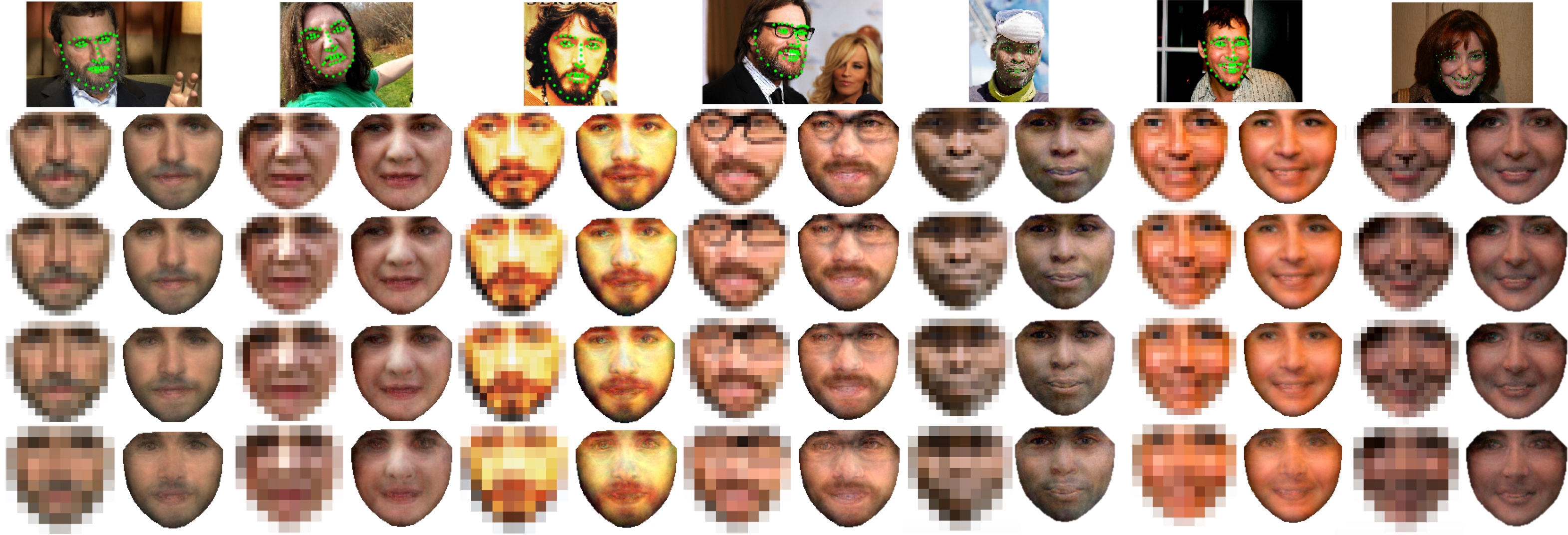}
	\end{center}
	\caption{Facial image super-resolution reconstruction at different scales of down-sampling. The 1st row: original image, the 2nd row to the 5th row: down-scaled images with factors of 4, 6, 8, 12 (left) and reconstructed facial images using DAMs (right). }
	\label{fig:lfwp_helen_super_res_scales}
\end{figure*}

\subsection{Properties of Deep Appearance Models} \label{subsec:DAMProp}

Deep Appearance Models provide the capability of generating facial shapes using texture information and vice versa. For example, one can predict a facial shape from the appearance using DAMs as follows: (1) clamping the texture information $\mathbf{g}$ as observations for the texture model and initializing hidden units with random states; (2) performing standard Gibbs sampling as a posterior inference step; and (3) obtaining the reconstructed shape from $P(\mathbf{s}|\mathbf{g};\theta)$.
To generate the appearance from a given shape, one can apply the same way with reversed pathways after clamping the shape information to the shape model. Figure \ref{fig:generated_shape_given_texture} represents the generated shapes given textures in three cases of expressions, occlusions and poses. In all these cases, the DAMs model is able to predict the shape correctly.

In addition, it is more natural to interpret both shapes and textures using higher hidden layers. In order to obtain this representation, one can clamp both observed shape $\mathbf{s}$ and texture $\mathbf{g}$ together before applying the Gibbs sampling procedure to estimate $P(\mathbf{h}^{(3)}|\mathbf{s},\mathbf{g};\theta)$. Eventually, probabilities of these hidden layers can be used as features. Notice that, besides the advantage of better features for discriminative tasks, one can easily see that even when one of two inputs is missing (i.e. shape), $P(\mathbf{h}^{(3)}|\mathbf{g};\theta)$ is still able to approximate. Hence, DAMs can be considered as a more generative model compared to other appearance models.

In terms of the number of training samples, with a specific topology of a stochastic neural network and sampling based weight update process (i.e. Contrastive Divergence), the need for large-scale training data is alleviated in our DAMs structure. Therefore, this is an advantage of our deep structure since it does not require a large-scale training dataset as other CNN based approaches.

The proposed method can also deal with facial reconstruction in various challenging conditions, such as: facial occlusions, facial expressions, facial off-angles, etc. These advantages of this method will be shown in Section \ref{sec:experiment}.

\section{Fitting in Deep Appearance Models} \label{sec:Model_fitting}
\subsection{Forward Composition Based Fitting} \label{sec:Forward_fitting}
Given a testing face $I$, the fitting process in DAMs can be formulated as finding an optimal shape $\mathbf{s}$ that maximizes the probability of the shape-free image as in Eqn. (\ref{eq:sfitting}).
\small
\begin{equation} \label{eq:sfitting}
\mathbf{s}^*=\arg\max_{\mathbf{s}}P(I(W(r_{\mathcal{D}},\mathbf{s}))|\mathbf{s};\theta)
\end{equation}
\normalsize
Since the connections between textures and hidden units $\mathbf{h}^{(1)}_g$ are modeled by a GRBM, the probability of texture $\mathbf{g}$ given hidden units $\mathbf{h}^{(1)}_g$ is computed as:
\small
\begin{equation}
P(\mathbf{g}|\mathbf{h}^{(1)}_g;\mathbf{s},\theta)=\mathcal{N}(\sigma_g\mathbf{W}^{(1)}_g\mathbf{h}^{(1)}_g+\mathbf{b}_g,\mathbf{\sigma}_g^2\mathbf{A})
\end{equation}
\normalsize
where $\mathbf{A}$ is the identity matrix; \{$\mathbf{\sigma}_g, \mathbf{b}_g$\} are the standard-deviation and bias of visible units in the texture model;  and $\mathbf{W}^{(1)}_g$ are learned weights of the visible-hidden texture. 

During the fitting steps, the states of hidden units $\mathbf{h}^{(1)}_g$ are estimated by clamping both the current shape $\mathbf{s}$ and the texture $\mathbf{g}$ to the model. The Gibbs sampling method is then applied to find the optimal estimated texture of the testing face given a current shape $\mathbf{s}$. By this way, the hidden units in DAMs can take into account both shape and texture information in order to reconstruct a better texture for further refinement.

Let $\mathbf{m}=\sigma_g\mathbf{W}^{(1)}_g\mathbf{h}^{(1)}_g+\mathbf{b}_g$ be the mean of the Gaussian distribution, we have the following approximation:
\begin{equation}
P(I(W(r_{\mathcal{D}},\mathbf{s}))|\mathbf{h}^{(1)}_g;\theta)=\mathcal{N}(\mathbf{m},\sigma_g^2\mathbf{A})
\end{equation}
The maximum likelihood can be then estimated as follows:
\small
\begin{equation}\label{AAMFitting}
\begin{split}
\mathbf{s}^*=&\arg\max_\mathbf{s}(P(I(W(r_{\mathcal{D}},\mathbf{s}))|\mathbf{s};\theta))\\
&=\arg\max_\mathbf{s}\mathcal{N}(I(W(r_{\mathcal{D}},\mathbf{s}))|\mathbf{m},\sigma_g^2\mathbf{A}))\\
&=\arg\min_\mathbf{s}\frac{1}{\sigma_g^2}\sum(I(W(r_{\mathcal{D}},\mathbf{s}))-\mathbf{m})^2
\end{split}
\end{equation}
\normalsize
Then the forward compositional algorithm can be used to solve the problem (\ref{AAMFitting}) by finding the updating parameter $\Delta \mathbf{s}$ that increases the likelihood:
\small
\begin{equation} \label{eq:Delta_shape_update}
\Delta \mathbf{s} = \arg\min_{\Delta \mathbf{s}}\|I(W(W(r_{\mathcal{D}},\Delta \mathbf{s}),\mathbf{s}))-\mathbf{m} \|^2
\end{equation}
\normalsize
The linearization is taken place of the test image coordinate using first order Taylor expansion $I(W(W(r_{\mathcal{D}},\Delta \mathbf{s}),\mathbf{s})) = I(W(r_{\mathcal{D}},\mathbf{s}))+\mathbf{J}_I\Delta \mathbf{s}$ and the update parameter is given as:
\small
\begin{equation}
\Delta \mathbf{s} = -(\mathbf{J}^T_I\mathbf{J}_I)^{-1}\mathbf{J}^T_I\left[I(W(r_{\mathcal{D}},\mathbf{s}))-\mathbf{m})\right]
\end{equation}
\normalsize
where $\mathbf{J}_I=\nabla I\frac{\partial W}{\partial \mathbf{s}}$ is the Jacobian.

\subsection{Dictionary Learning Based Fitting}
In this section, we further improve the fitting process so that it can deal with occlusions and other variations.
From Eqn. (\ref{eq:Delta_shape_update}), we can see that the shape update $\Delta \mathbf{s}$ mostly relies on the difference between the shape-free image and its DAMs reconstruction. However, this metric is easily affected by the presence of occlusions. In particular,
when part of the face is occluded, the occlusion still remained in the shape-free image as the result of warping operator but will be removed in DAMs reconstruction. This will lead to the cases where the $\ell_2$ distance between the two images is still very large even the optimization approaches the correct shape. As a result,  it will mislead the optimization process and the final shape cannot be optimized.
Therefore, the $\ell_2$-norm of their difference is not robust enough to guide the fitting process to the true shape when occlusions occur.

To address this problem more effectively, instead of working directly in texture space, we define a mapping function $f:\mathcal{I} \mapsto \mathcal{C}$ to map $I(W(r_{\mathcal{D}},\mathbf{s}))$ and $\mathbf{m}$ from texture space $\mathcal{I}$ to a parameter space $\mathcal{C}$ such that the relationship between $f(I(W(r_{\mathcal{D}},\mathbf{s})))$ and $f(\mathbf{m})$ is more robust to occlusions. Then this relationship can be used for fitting process.
The mapping function $f$ can be defined as
\begin{equation}
\begin{split}
f:&\mathcal{I} \mapsto \mathcal{C}\\
& \mathbf{c}_1 = f(I(W(r_{\mathcal{D}},\mathbf{s}))) \\
& \mathbf{c}_2 = f(\mathbf{m})\\
\end{split}
\end{equation}
Then it can be parameterized by dictionaries and representation coefficients as follows.
\small
\begin{eqnarray} \label{eq:Dict_coeff_estimate}
\begin{aligned}
f(I(W(r_{\mathcal{D}},\mathbf{s}))) = \arg \min_{\mathbf{c}_1} & \parallel I(W(r_{\mathcal{D}},\mathbf{s})) - \mathbf{\hat{D}}_I \mathbf{c}_1\parallel^2_2 \\
& + \lambda_1 \parallel \mathbf{c}_1\parallel_1 \\
f(\mathbf{m}) = \arg \min_{\mathbf{c}_2} & \parallel \mathbf{m} - \mathbf{\hat{D}_m} \mathbf{c}_2\parallel^2_2 + \lambda_2 \parallel \mathbf{c}_2\parallel_1
\end{aligned}
\end{eqnarray}
\normalsize
where $\{\mathbf{\mathbf{\hat{D}}}_I,\mathbf{c}_1\}$ and $\{\mathbf{\hat{D}}_\mathbf{m}, \mathbf{c}_2\}$ are the dictionaries and representation coefficients of the shape free image and its DAMs reconstruction, respectively. 
The DAMs fitting can be decomposed into two steps, i.e. training and testing.

\textbf{Training step}: Given a training dataset with $N$ images and their shapes $\{(I^i,\mathbf{s}^i)\}_{i=1}^N$, the dictionaries are learned by minimizing 
the objective function
\small
\begin{eqnarray} \label{eq:Dict_combined}
\begin{aligned}
\{\mathbf{\hat{D}}_I,\mathbf{\hat{D}_m}\} = \arg \min_{\mathbf{D}_I,\mathbf{D_m} \in \mathbb{R}^{k \times l}}\frac{1}{N}\sum_{i=1}^{N} \{
&\min_{ \mathbf{c}^i\in \mathbb{R}^l}\parallel \mathbf{I}_W^i - \mathbf{D}_I \mathbf{c}^i\parallel ^2_2 \\ 
&+ \parallel \mathbf{m}^i - \mathbf{D_m} \mathbf{c}^i\parallel ^2_2 \\
&+ \lambda \parallel \mathbf{c}^i\parallel_1
\}
\end{aligned}
\end{eqnarray}
\normalsize
where $\mathbf{I}_W^i = I^i(W(r_{\mathcal{D}},\mathbf{s}))$, $k$ is the length of texture vector and $l$ is the size of dictionaries.
With this objective function, the two dictionaries $\mathbf{\mathbf{\hat{D}}}_I$ and $\mathbf{\hat{D}}_\mathbf{m}$ are learned in a way that regardless of the present of occlusions in the input face, the extracted representation coefficients of $I(W(r_{\mathcal{D}},\mathbf{s}))$ and $\mathbf{m}$ are forced to share the same (i.e. $\mathbf{c} = \mathbf{c}_1 = \mathbf{c}_2$) when the true shape is approaching.
To solve \eqref{eq:Dict_combined}, we apply the four-step iterative procedure as in ~\citep{xing2014towards}. The main steps of this procedure are summarized in Algorithm \ref{Alg:Dictionary_learning}. There are two main advantages of learning the dictionaries as in Eqn. (\ref{eq:Dict_combined}). Firstly, since both shape-free image and its DAMs reconstruction are forced to share the same representation $\mathbf{c}^i$, their underlying relationships are naturally embedded in these coefficients. Secondly, when the vector $\mathbf{c}^i$ is sparse, the optimization will result in the most related features between the shape-free image and its reconstruction. Therefore, it will be more robust to occlusions and other variations.
\begin{figure*}[t]
	\begin{center}
		\includegraphics[width=14.5cm]{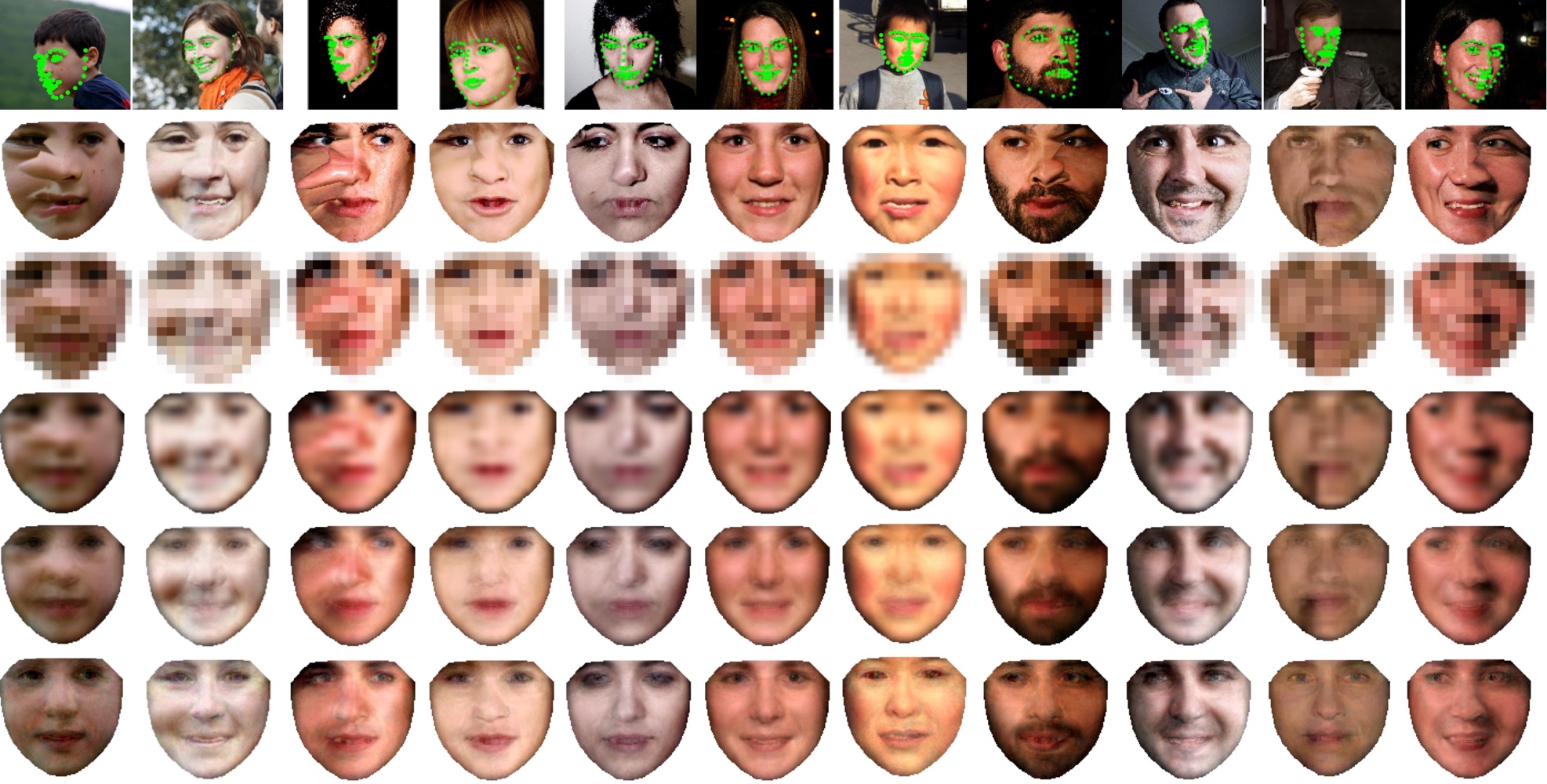}
	\end{center}
	\caption{Facial image super-resolution. The original images (first row) are warped to shape-free images in texture domain (second row); then they are down-sampled by a factor of 8 from $117 \times 120 $ to $15 \times 15$ (third row)
		The next three rows are the high-resolution reconstructed using Bicubic method (the fourth row), PCA-based AAMs (the fifth row) and Deep Appearance Models (the sixth row).}
	\label{fig:lfwp_helen_superresolution}
\end{figure*}

\begin{algorithm}
	\caption{Dictionary Learning for fitting}\label{Alg:Dictionary_learning}
	\begin{algorithmic}[1]
		\Require Training data $\{(I^i,\mathbf{s}^i)\}_{i=1}^N$, regularization parameter $\lambda$
		\Ensure Learned dictionaries $\{\mathbf{\hat{D}}_I,\mathbf{\hat{D}_m}\}$
		\State Construct the matrix $\mathbf{Y}\in \mathbb{R}^{k \times N}$ whose $i$-th column is shape-free image $\mathbf{I}_W^i$.
		\State Construct the matrix $\mathbf{M}\in \mathbb{R}^{k \times N}$ whose $i$-th column is $\mathbf{m}^i$.
		\State Initialize $\mathbf{D}_I  \in \mathbb{R}^{k \times l}$ and $\mathbf{D_m}  \in \mathbb{R}^{k \times l}$ with random samples from a normal distribution with zero mean and unit variance.
		\While{not converged}
		\State (1) Fix $\mathbf{D_m}$, learn $\mathbf{D}_I$ and coefficient matrix $\mathbf{C} \in \mathbb{R}^{l \times N}$
		\small
		\begin{equation}
		\{\mathbf{D}_I, \mathbf{C}\} = \arg \min_{\mathbf{D}_I,\mathbf{C}}
		\parallel \mathbf{Y} - \mathbf{D}_I \mathbf{C}\parallel ^2_2
		+ \lambda \parallel \mathbf{C}\parallel_1 \nonumber
		\end{equation}
		\normalsize
		\State (2) Update $\mathbf{D_m}$ as $\mathbf{D_m} = \mathbf{M/C}$.
		Notice that this result is used as initial $\mathbf{D_m}$ for step (3).
		\State (3) Fix $\mathbf{D}_I$, learn $\mathbf{D_m}$ and new coefficient matrix $\mathbf{C}$
		\small
		\begin{equation}
		\{\mathbf{D_m}, \mathbf{C}\} = \arg \min_{\mathbf{D_m},\mathbf{C}}
		\parallel \mathbf{M} - \mathbf{D_m} \mathbf{C}\parallel ^2_2
		+ \lambda \parallel \mathbf{C}\parallel_1 \nonumber
		\end{equation}
		\normalsize
		\State (4) Update $\mathbf{D}_I$ as $\mathbf{D}_I = \mathbf{Y/C}$.
		\EndWhile
		\State Set $\mathbf{\hat{D}}_I = \mathbf{D}_I$ and $\mathbf{\hat{D}_m} = \mathbf{D_m}$.
	\end{algorithmic}
\end{algorithm}

After obtaining the dictionaries, instead of following the Gaussian-Newton optimization as in Eqn. (\ref{eq:Delta_shape_update}), we learn a linear regressor to directly infer the shape update $\Delta \mathbf{s}$ from the difference between $f(I_W)$ and $f(\mathbf{m})$. Specifically, given training images with their initial estimate $\{\mathbf{\bar{s}}^i\}_{i=1}^N$ of their ground truth shape, the linear regressor is learned by minimizing
\begin{equation}
\arg\min_{\mathbf{H,b}}\sum_{i=1}^{N} \parallel \Delta \mathbf{s}^i -\mathbf{H}\left[f(I_W^i)-f(\mathbf{m}^i)\right] - \mathbf{b}\parallel^2
\end{equation}
where $\Delta \mathbf{s}^i = \mathbf{s}^i - \mathbf{\bar{s}}^i$; $\{(\mathbf{H,b})\}$ are the regressor's parameters.

\textbf{Testing step}: In the fitting process, given an input face with its initial shape, the shape-free image $I_W$ and DAMs reconstruction $\mathbf{m}$ are first computed. Their representation coefficients $\mathbf{c}_1^*$ and $\mathbf{c}_2^*$ are also estimated using Eqn. (\ref{eq:Dict_coeff_estimate}). Then the shape is updated using the difference $\mathbf{c}_1^* - \mathbf{c}_2^*$ together with the learned regressor. After that the image pair $(I_W, m)$ is recomputed for the next iteration.

\section{Experimental Results} \label{sec:experiment}

In section \ref{subsec:databases}, we briefly introduce the main features of the three databases used in our evaluations. They consist of two ``face in the wild'' and one aging databases. By using these databases with numerous challenging factors, we aim to show the robustness and efficiency of our proposed model.
Then, in the next three sections, we validate the generative capabilities of our DAMs  in both facial representation and reconstruction via four applications, i.e. facial super-resolution, facial off-angle reconstruction, facial occlusion removal, and facial age estimation. The experiments are also made to be more challenging by including numerous variations in poses, occlusions and impulsive noise. Comparing to AAMs, bicubic interpolation and other deep learning based approaches, our DAMs achieve better reconstructions without blurring effects or spreading out the errors caused by occlusions or noise.
We also represent in section \ref{subsec:Shape_fitting} an experiment to evaluate our proposed DAMs method in the ability of synthesizing new face images. Its performance 
is 
compared with AAMs and other face alignment methods 
such as RCPR ~\citep{burgos2013robust}, and CNN based approach ~\citep{sun2013deep}.

\subsection{Databases} \label{subsec:databases}
\label{subsec:db}
We aim to build a model that can represent face texture in-the-wild. Therefore, in the first three applications, we evaluate DAMs on two face databases in-the-wild, i.e. Labeled Face Parts in the Wild (LFPW) ~\citep{belhumeur2011localizing} and Helen ~\citep{le2012interactive}.
These databases contain unconstrained facial images collected from various multimedia resources. These facial images have considerable resolutions and contain numerous variations such as poses, occlusions and expressions. For the age estimation application, FG-NET face aging database\footnote{The FG-NET Aging Database, http://www.fgnet.rsunit.com/.} 
is used to evaluate the method.

The LFPW database contains 1400 images in total with 1100 training and 300 testing images. However, a part of it is no longer accessible. Therefore, in our experiments, we only use 811 training and 224 testing images, the available remaining. Each facial image is annotated with 68 landmark points provided by 300-W competition ~\citep{sagonas2013semi}.

The Helen database provides a high-resolution dataset with 2000 images used for training and 330 images for testing. The variations consist of pose changing from $-30^\circ $ to $30^\circ$; several types of expression such as neutral, surprise, smile, scream; and occlusions. Similar to LFPW, all faces in Helen are also annotated with 68 landmark points.

FG-NET is a popular face aging database. There are 1002 face images of 82 subjects with age ranges from 0 to 69 years. The annotations in FG-NET are also 68 landmarks in the same format as LFPW and Helen databases.

\begin{figure*}[!t]
	\begin{center}
		\includegraphics[width=15cm]{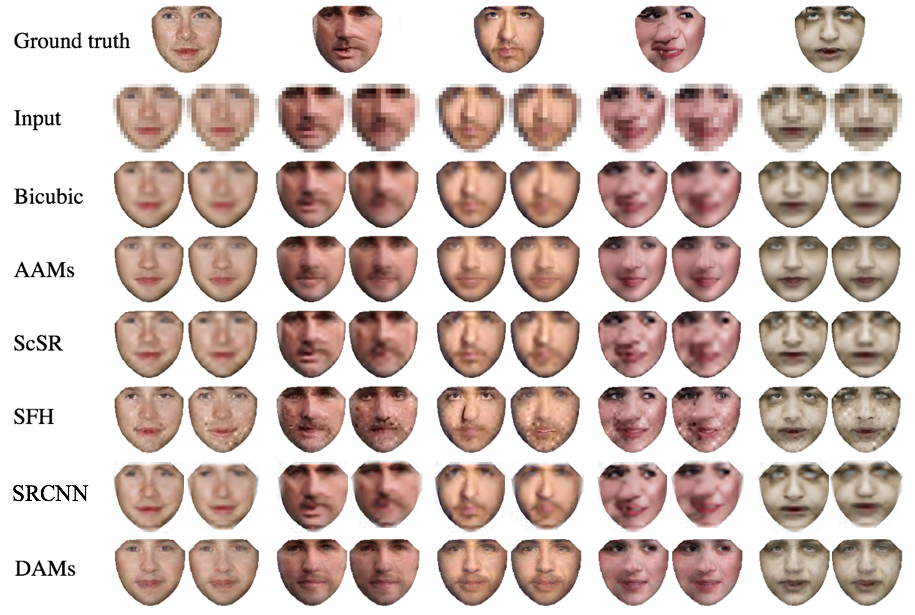}\end{center}
	\caption{Comparisons of different facial image super-resolution methods. The 1st row: ground truth faces. The 2nd row: down-scaled images with factors of 6 (left) and 8 (right). From the 3rd row to the 8th row: reconstructed faces using bicubic, PCA-based AAMs, ScSR ~\citep{yang2010image}, SFH ~\citep{yang2013structured}, SRCNN ~\citep{dong2014learning} and DAMs, respectively.}
	\label{fig:SR_compare_new_all}
\end{figure*}
\subsection{Facial Super-resolution Reconstruction} \label{subsec:face_super_res}

The proposed DAMs method is evaluated in its capability to recover high-resolution face images given their very low-resolution versions.
Moreover, since LFPW and Helen data-bases also include numerous variations in poses, expressions and occlusions, the experiment becomes more challenging. Our proposed method is very potential in dealing with the problem of super-resolution in various conditions of facial poses and occlusions.

In order to train the DAMs model, we combine 811 training images from LFPW and 2000 images from Helen data-base into one training set.
The coordinates of facial landmarks are normalized to zero mean before setting as observations to train the shape model.
In the texture modeling, shape-free images are first extracted by warping faces into the texture domain $\mathcal{D}$. The size of the shape-free image is set to $117 \times 120$ pixels based on the mean shape of the training data. Then texture model is trained to learn the facial variations represented in these shape-free images.

During the testing phase, since the number of visible units in the texture model is fixed, the testing low-scale facial shape-free image is first resized to $117 \times 120$ using bicubic interpolation method. Then both the shape and the shape-free image are clamped to DAMs. After 50 epochs in the alternating Gibbs updates, the face texture is reconstructed based on the current states of hidden unit $\mathbf{h}^{(1)}_g$.
Different magnification factors $\alpha$ are used for evaluating the quality of DAMs reconstructions. Testing images are down-sampled in  different magnification levels ranging from 4 to 12. They are then used as inputs to the reconstruction module of our approach.
Figure \ref{fig:lfwp_helen_super_res_scales} shows the reconstructions using the DAMs approach. Remarkable results are achieved using DAMs with very low-resolution input images, i.e. $10 \times 10$ pixels with the magnification factor $\alpha = 12$.

\textbf{Comparisons against Baseline Methods}: Our proposed approach is also compared with two base-line methods, i.e. bicubic interpolation method and PCA-based AAMs ~\citep{tzimiropoulos2013optimization}.
~Root Mean Square Error (RMSE) is used as a performance measurement. RMSE is a common metric that is usually used for evaluating image recovery task. Although this metric is not always reliable for rating image quality visually ~\citep{wang2009mean}, it could provide a qualitative view for comparing DAMs and other methods.

\begin{figure}[!t]
	\begin{center}
		\includegraphics[width=6.5cm]{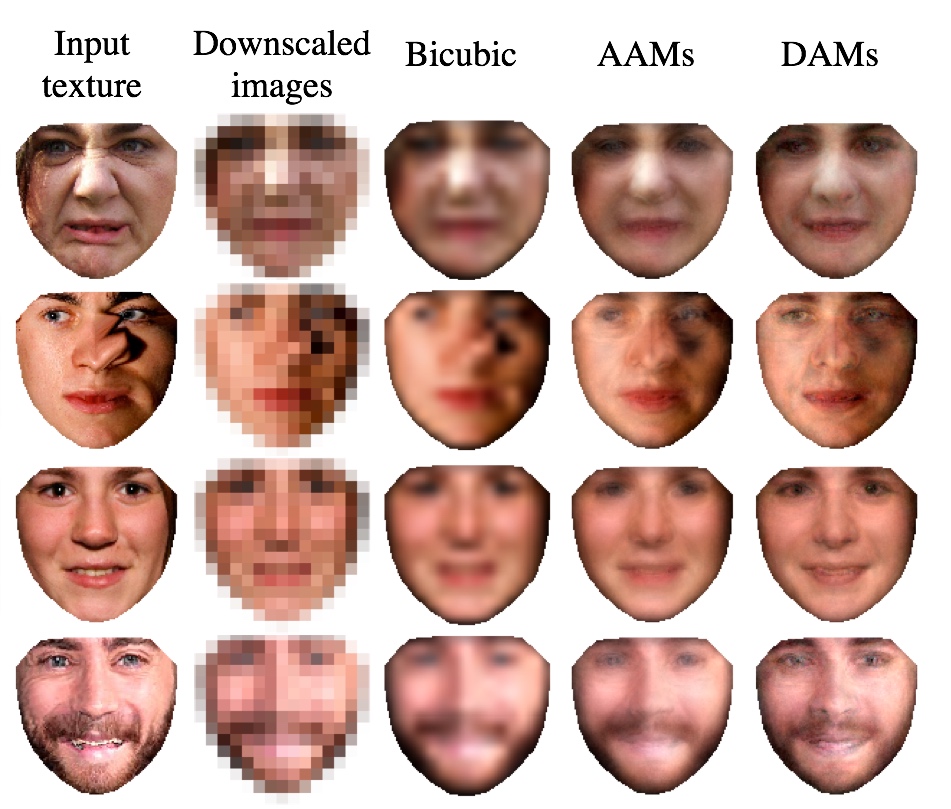}\end{center}
	\caption{Results of average RMSEs over 4 images: Bicubic interpolation (RMSE = 19.68); PCA-based AAMs reconstruction (RMSE = 19.96); (d) Deep Appearance Models reconstruction (RMSE = 20.44).}
	\label{fig:RMSE_not_good}
\end{figure}

\begin{table}[t]
	\caption{The average RMSEs of reconstructed images using different methods against LFPW and Helen databases with $\alpha = 16$.}
	\label{table_RMSE_super_res_combined}
	\centering
	\begin{tabular}{|c|c|c|}
		\hline
		\textbf{Methods} & \textbf{LFPW} & \textbf{Helen} \\
		\hline
		Bicubic & 19.53 & 22.13  \\
		\hline
		AAMs & 19.74 & 22.3 \\
		\hline
		DAMs (Ours) & \textbf{19.24} & \textbf{21.24}\\
		\hline
	\end{tabular}
\end{table}

\begin{table}[!t]
	\caption{The average PSNRs (dB) of different methods on LFPW and Helen.}
	\label{table_PSNR_superres}
	\centering
	\begin{tabular}{|c|c|c|c|c|c|c|}
		\hline
		\multirow{2}{*}{\textbf{Methods}} & \multicolumn{2}{c|}{$\alpha=4$} &  \multicolumn{2}{c|}{$\alpha=6$} &  \multicolumn{2}{c|}{$\alpha=8$}\\
		\cline{2-7}
		& LFPW & Helen & LFPW & Helen & LFPW & Helen\\
		\hline
		Bicubic & 19.47 & 19.22 & 18.82 &  18.59 & 18.29 & 18.07\\
		\hline
		AAMs & 26.43 & \textbf{26.56} & 25.72 & 25.83 & 24.82 & 24.97  \\
		\hline
		SRCNN  & \textbf{26.73} & 26.46 & 24.61 & 24.30 & 23.22 & 22.85\\
		\hline
		DAMs & 26.46 & 26.54 & \textbf{25.91} & \textbf{25.97} & \textbf{25.06} & \textbf{25.16}\\
		\hline
	\end{tabular}
\end{table}
From the results shown in Figure \ref{fig:RMSE_not_good}, our method gives better reconstruction results in visualization than the others. However, the RMSE results are not much better as shown in Table \ref{table_RMSE_super_res_combined}. This is because RMSE cannot fully evaluate the quality of reconstructed images in the task of image super-resolution ~\citep{yang2010image}. Especially, we don't have the ground-truth for RMSE evaluation in these databases.
For example, in the cases of occlusions and poses in those databases, although the reconstructed images obtained using PCA-based AAMs and bicubic methods are very blurry, their RMSEs are still low. This is because
the reconstructed images still contain occlusion components or pose features  which are quite similar to the original ones.
Figure \ref{fig:lfwp_helen_superresolution} illustrates further reconstruction results obtained using bicubic method, PCA-based AAMs method and our DAMs approach. The PCA-based AAMs method is trained using the same dataset as DAMs and the length of texture parameter vector is 200, the highest level used in ~\citep{tzimiropoulos2013optimization}).

\begin{figure}[!t]
	\begin{center}
		\includegraphics[width=8.5cm]{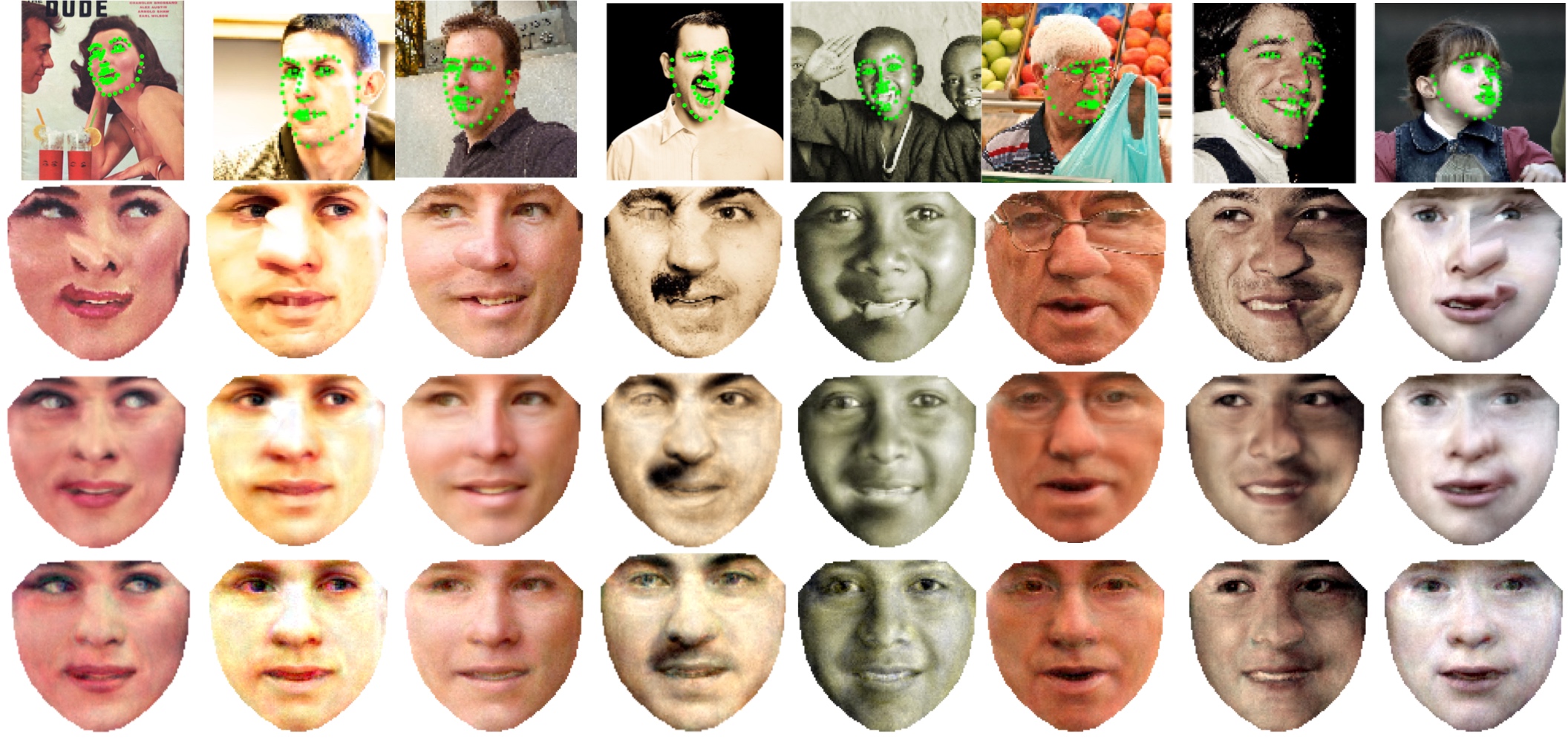}\end{center}
	\caption{Facial off-angle reconstruction: the 1st row: original image, the 2nd row: shape-free image, the 3rd row: PCA-based AAMs reconstruction 
		, and the 4th-row: DAMs reconstruction}
	\label{fig:headpose_reconstruct}
\end{figure}

\textbf{Comparisons against Other Super-resolution Methods}:
For further evaluations, we  compare DAMs with three other super-resolution methods in Figure \ref{fig:SR_compare_new_all}. The other three super-resolution methods are: sparse representation based image super-resolution (ScSR) \citep{yang2010image}, ~Structured Face Hallucination (SFH) \citep{yang2013structured}, ~and super-resolution CNN (SRCNN) based approach \citep{dong2014learning}. The main difference between the first two is that the former is designed for images in general while the latter is more specific for facial images. The third approach is a deep learning based method. 
For each subject, the low-resolution (LR) faces (i.e. $LR\_6$ and $LR\_8$) are obtained by down-sampling the ground truth face with factors of 6 and 8. Their high-resolution (HR) reconstructed faces (i.e. $HR\_6$ and $HR\_8$) of different methods are shown in the left and right columns underneath each ground truth face in the first row, respectively. The results of Bicubic and AAMs are also presented in this figure. The resolution of $LR\_6$ is $20 \times 20$ and that of $LR\_8$ is $15 \times 15$.

It is clear in the figure that SFH performs better than ScSR in terms of reconstruction details. This is because SFH was already trained with the face structure and contour's statistical priors.
However, some noisy and blocky effects are still remained in the reconstructed faces of SFH. Especially, when parts of face images are blurred due to the effects of warping operator, artifacts may appear in the final results. The SRCNN also shows some advantages with small magnification factor  $\alpha$ but the blurry effects are still presented when $\alpha$ increases.
Meanwhile, remarkable results can be achieved by DAMs in terms of keeping fine details without noisy effects.
In addition, these results also show the advantages of DAMs when dealing with higher magnification factor $\alpha$.
Whereas all five methods fail to produce high quality reconstructions when $\alpha$ increases from 6 to 8, DAMs still perform well and generate faces with consistent quality.

Table \ref{table_PSNR_superres} presents another qualitative comparison in terms of average PSNR values between our DAMs against one deep learning based  (i.e. SRCNN) and two baseline methods. These results again show the effectiveness of DAMs in this task. With small value of $\alpha$, DAMs produce comparable results to AAMs and SRCNN. When $\alpha$ increases (i.e. from 4 to 8), DAMs achieve the highest PSNRs as compared to other methods.

\subsection{Facial off-angle Reconstruction and Occlusion Removal} \label{subsec:face_off_res}
This section illustrates the ability of DAMs to deal with facial poses and occlusions.

\subsubsection{Facial off-angle Reconstruction}

Using the same trained model as in the previous experiment, facial images with different poses are represented in Figure \ref{fig:headpose_reconstruct}.
Comparing to AAMs, our DAMs achieve better reconstructions especially in the invisible regions of extreme poses.
These regions in shape-free images are blurry and noisy due to the non-linear warping operator. Therefore, the errors are spread out in the reconstructions of PCA-based AAMs approaches.
Meanwhile, the generative capability of our proposed DAMs method can solve those challenging cases. From the results, it is easy to see that the blurry effects are effectively removed in DAMs reconstructions.

\begin{figure*}[!t]
	\begin{center}
		\includegraphics[width=17cm]{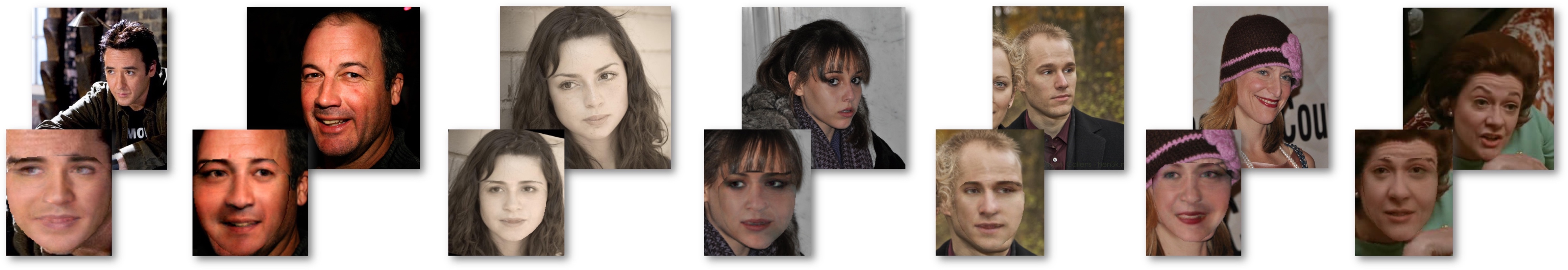}\end{center}
	\caption{Face Frontalization: Top: input faces and Bottom: frontalized faces reconstructed using DAMs.}
	\label{fig:face_frontalization}
\end{figure*}

\begin{figure*}[t]
	\begin{center}
		\includegraphics[width=16.5cm]{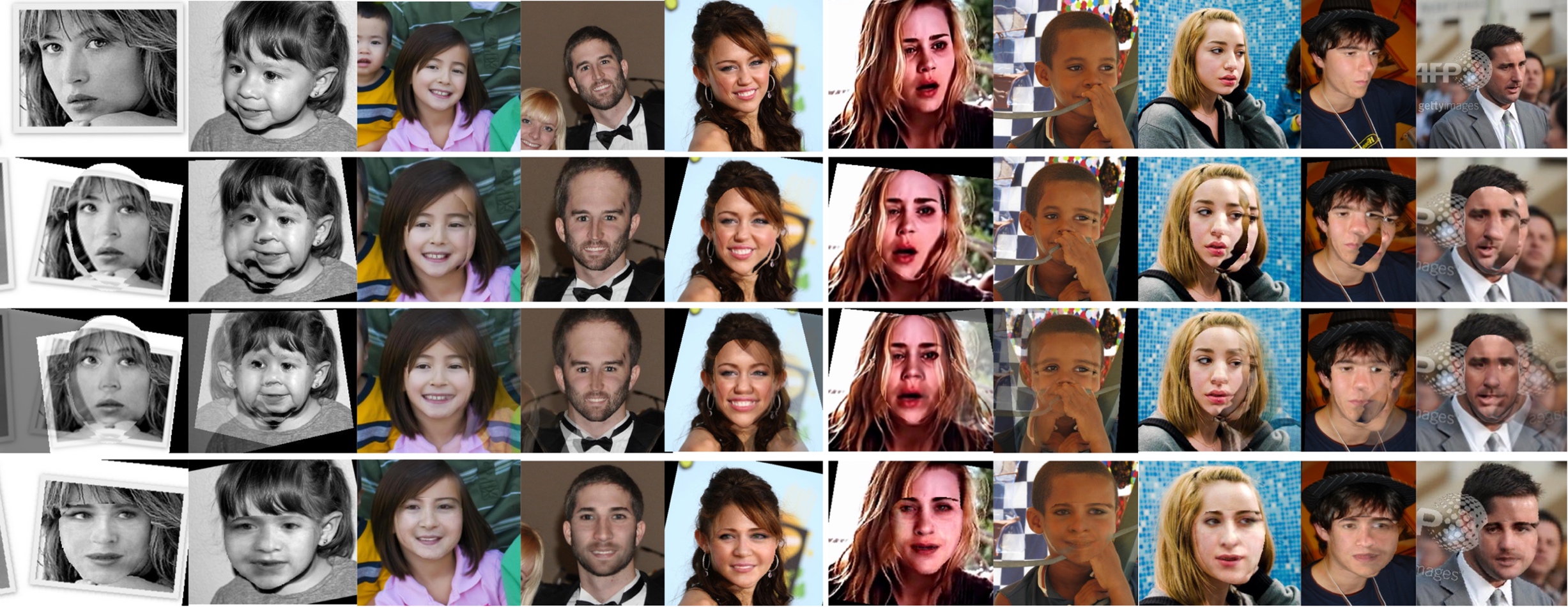}\end{center}
	\caption{Comparisons between DAMs and Face Frontalization approach ~\citep{HHPE:CVPR15:frontalize}. The 1st row: input faces; the 2nd and 3rd rows: synthesized frontal view before and after applying soft symmetry ~\citep{HHPE:CVPR15:frontalize}; the 4th row: frontalized faces produced by DAMs. }
	\label{fig:face_frontalization_compare}
\end{figure*}

\subsubsection{Face Frontalization}
We next emphasize this ability of our DAMs approach on the face frontalization problem.
Given an input face with pose, the process of ``frontalization'' is to synthesize the frontal view of that face. Notice that the facial photos are unconstrained and the subjects are not required to already be in the training data. Once again, in order to produce aesthetic frontal view, not only poses but other factors such as expressions and occlusions are needed to be taken into account.
The frontalization can help to boost the performance of other subsequent processes such as face recognition, verification, 
gender estimation 
~\citep{HHPE:CVPR15:frontalize}, etc.
There are several frontalization approaches proposed in literature ~\citep{sagonas2015robust, HHPE:CVPR15:frontalize, wang2015robust, jeni2016person, ferrari2016effective}. \cite{sagonas2015robust} employed a statistical model built on some frontal images to simultaneously reconstruct frontal view and align faces. Frontal and aligned faces are then obtained by solving a rank minimization problem. Inspired by the use of 3D model for frontal face reconstruction, \cite{HHPE:CVPR15:frontalize} proposed to approximate facial shapes using a single 3D surface while \cite{wang2015robust} tried to fit the 3D model to the input 2D face by iteratively refining the 3D landmarks and the weighting coefficients of each landmark. \cite{jeni2016person} later introduced a cascade regression-based approach to estimate a canonical view of the eyes for 3D gaze estimation. The algorithm includes several steps: localizing a set of dense landmarks, fitting a part-based 3D model for 3D shape reconstruction and estimating head pose and gaze. \cite{ferrari2016effective} proposed to use a 3DMM to fit the input image and then map each image pixel to its 3D corresponding coordinate of the model to obtain a frontal view. By projecting the 3D model back to the frontalized image, image patches can be located and aligned for feature extraction over different images.

Figure \ref{fig:face_frontalization} represents the frontalized views of input faces with different poses and expressions given in the top row.
\begin{table}[t]
	\caption{The face verification accuracies on LFW benchmark.}
	\label{table_FaceRecog_LFW}
	\centering
	\begin{tabular}{|c|c|c|c|}
		\hline
		\textbf{Methods} & \textbf{Original inputs} 
		& \textbf{LFW3D} 
		& \textbf{DAMs}\\
		\hline
		Pittpatt Classifier & 83.67\% & 86.63\%& \textbf{87.72\%}\\
		\hline
	\end{tabular}
\end{table}
Our reconstruction results are also compared with the recent frontalization work ~\citep{HHPE:CVPR15:frontalize} against LFPW and Helen databases in Figure \ref{fig:face_frontalization_compare}. From the second and third rows, one can see that the approach in ~\citep{HHPE:CVPR15:frontalize} achieves good reconstructions when the input poses are not so extreme (i.e. not greater than 30 degrees).
However, in case of extreme poses (i.e. the first two and the last three faces) or occlusions (the 7th face), even when the symmetry property is used, the full faces can not be reconstructed aesthetically.
Meanwhile, the results in the last row show that DAMs can effectively synthesize the frontal views of these faces without further applying the soft symmetry property. Since the face priors are already learned, DAMs are able to produce more natural faces instead of duplicating the information from known side to the other side. To further compare the effectiveness of DAMs against this frontalization work (i.e. LFW3D), we also employ the face verification protocol on the Labeled Face in the Wild (LFW) dataset and achieve the results as in Table \ref{table_FaceRecog_LFW}. Noticed that applying directly deep learning based classifiers may mask out the contributions of frontalization step, we employ the off-the-shelf commercial face recognition Pittpatt (developed by CMU and Google) in this evaluation. These results show that our DAMs can handle the face poses effectively and produce a higher accuracy boost comparing to LFW3D approach.

There are some other deep learning based frontalization approach such as FIP ~\citep{zhu2013deep} and MVP ~\citep{zhu2014multi}. However, both approaches require several images of the same identity with different poses to train the deep models. Moreover, other variations such as expressions, illuminations, and occlusions are kept unchanged between the input and output images during the learning stage.
These requirements have limited the use of these models with ``in-the-wild'' databases where each subject has only one image.
Meanwhile, our DAMs approach provides a model structure that is able to handle many variations at the same time without requirements on the number of images per subject as well as the view labels during training stage.

\begin{figure}[t]
	\begin{center}
		\includegraphics[width=8.5cm]{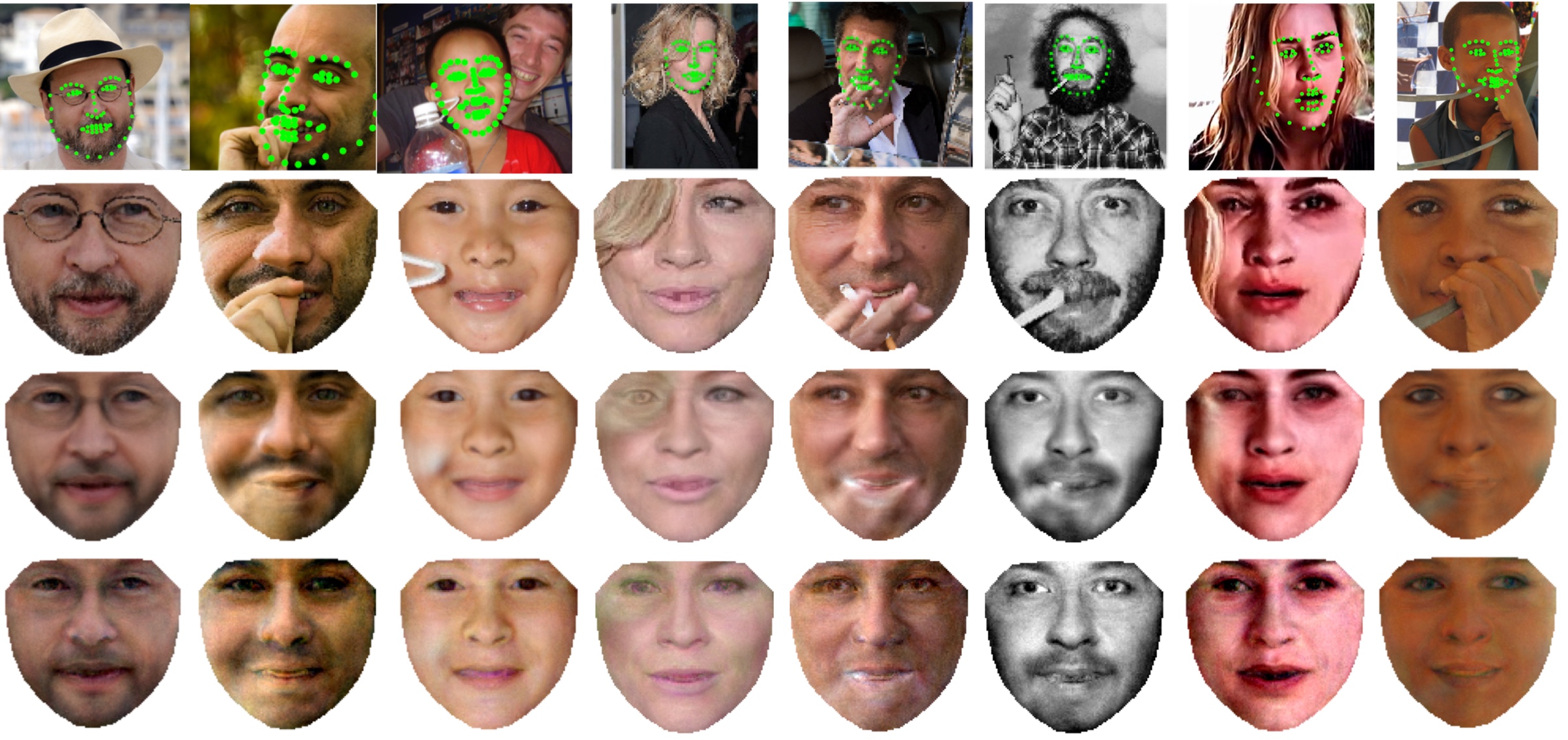}\end{center}
	\caption{Occlusion removal: the 1st row: original image, the 2nd row: shape-free image, the 3rd row: PCA-based AAMs reconstruction still remains with occlusion and blurring effects, and the 4th-row: DAMs reconstruction can help to remove the occlusion}
	\label{fig:occlusion_removal}
\end{figure}
\subsubsection{Facial Occlusion Removal}
Similarly, DAMs also show their capability in the problem of facial occlusion removal. In Figure \ref{fig:occlusion_removal}, the occlusions, e.g. hands, glasses, hair, etc.,  can be removed successfully without blurring effects. More interestingly, the occlusions are removed from faces without loosing facial features. For example, glasses are totally removed without making beard blurred as in  AAMs reconstruction.

Using occluded faces as references and measuring the reconstruction quality by RMSE cannot illustrate the modeling capabilities of DAMs.
To get a better evaluation protocol, we select a subset of 174 occluded faces of the first 29 subjects, i.e. 15 males and 14 females, from AR database ~\citep{martinez1998ar}. We employ DAMs to reconstruct these occluded faces and then use their corresponding neutral faces, i.e. frontal face without occlusions, as references to compute the RMSE.
In this testing set, each subject includes two faces with scarf and four other faces with both illumination and scarf. The average RMSE of DAMs is 45.08 while that of PCA-based AAMs is 47.36.
The trained models in DAMs and AAMs use LFPW and Helen databases as presented in Section \ref{subsec:db}.
This experiment shows that DAMs achieve better reconstructions, i.e. closer to the neutral faces, compared to AAMs.

\begin{table}[t]
	\caption{The rank-1 recognition results (\%) on the AR database.}
	\label{table_FaceRecog_AR}
	\centering
	\begin{tabular}{|c|c|c|c|c|}
		\hline
		\textbf{Methods} & \textbf{Pixels} & \textbf{AAMs} 
		& \textbf{RoBM} 
		& \textbf{DAMs}\\
		\hline
		Sunglasses & 51.24\% & 69.40\% & 71.89\%& \textbf{76.87\%}\\
		\hline
		Scarf & 55.97\% & 61.44\% & 64.18\% & \textbf{70.41\%}\\
		\hline
	\end{tabular}
\end{table}

In order to further illustrate the effectiveness of DAMs in the problem of facial occlusion removal, we also compare our DAMs with Robust Boltzmann Machines (RoBM) ~\citep{tang2012robust} in terms of recognition performance on AR.
We set up a similar protocol as in ~\citep{tang2012robust}. However, in that protocol, occluded faces of all subjects are also included in the gallery set. This may easily cause a ``match'' between occluded faces of the same subject in the gallery and probe sets. Therefore, in this experiment, we use only the non-occluded faces (i.e. seven images per subject) to compose the gallery set and leave all occluded faces (i.e. three with sunglasses and three with scarf per subject) for the probe set.
The recognition using DAMs consists of first reconstructing the ``clean'' faces using DAMs followed by a classification based on LDA and the nearest neighbor classifier. The cosine distance is used for the matching score.
Table \ref{table_FaceRecog_AR} shows the rank-1 recognition results obtained by different models. DAMs outperform other methods and illustrate their effectiveness in handling the face occlusions.

\subsection{Facial Age Estimation}

Besides some other previous age estimation approaches ~\citep{fu2008human,luu2009age}, we employ our proposed DAMs to this problem to further demonstrate their robustness and effectiveness.

\textbf{Evaluation on Reconstructed Images:}
Since texture is an important factor to predict a person's age given his facial image, this experiment will evaluate how good the reconstructed image is as well as how much aging information is retained by the model.

To make this task more challenging, we add noise to the testing facial image and then predict the age of that person using ``clean" reconstructed face from DAMs. 
For the evaluation system, we modified the age estimation systems presented in ~\citep{luu2009age, luu2011ageFG, luu2010ageCVPRW, luu2011ageIJCB, Duong2011ageICASSP} with three-group classification in the first step (youths, adults, and elders) before constructing three Support Vector Regression (SVR) based aging functions. Then we train this age estimator with 802 images from FG-NET. The remaining 200 images were used for testing.
To generate noisy testing images, all pixels of facial images were mixed with uniform noise ranged within $[-r, r]$.

A similar experiment is set up as follows: given the low-resolution testing face, the system will predict the age of that person using his high-resolution reconstructed face.
The Mean Absolute Errors (MAEs) of different methods against noise and low-resolution testing faces are represented in Table \ref{table_MAE_age_estimation} and Table \ref{table_MAE_age_estimation_super}, respectively.
From these results, in both cases, the smallest error is achieved with DAMs model.
Therefore, our proposed model produces better reconstructed results under the effects of noise and low-resolution factor.

\begin{table}[!t]
	\caption{The MAEs (years) of different methods against impulsive noise.}
	\label{table_MAE_age_estimation}
	\centering
	\begin{tabular}{|c|c|c|c|c|c|}
		\hline
		\multirow{2}{*}{\textbf{Methods}} &\multirow{2}{*}{\textbf{No noise}} & \multicolumn{4}{|c|}{\textbf{Noise range}} \\
		\cline{3-6}
		&  &  25 & 50 & 100 & 150\\
		\hline
		AAMs 
		& 6.14 & 6.15 & 6.11 & \textbf{6.13} & 6.47 \\
		\hline
		DAMs  & \textbf{5.67} & \textbf{5.81} & \textbf{5.56} & 6.14 & \textbf{6.18} \\
		\hline
	\end{tabular}
\end{table}
\begin{table}[!t]
	\caption{The MAEs (years) of different methods against low-resolution testing faces.}
	\label{table_MAE_age_estimation_super}
	\centering
	\begin{tabular}{|c|c|c|c|c|}
		\hline
		\multirow{2}{*}{\textbf{Methods}} & \multicolumn{4}{c|}{\textbf{Magnification factor $\alpha$}} \\
		\cline{2-5}
		& 2 & 4 & 6 & 8\\
		\hline
		Bicubic & 5.96 & 6.95 & 7.15 &  7.21\\
		\hline
		AAMs 
		& 6.13 & 6.33 & 6.44 & 6.69  \\
		\hline
		DAMs & \textbf{5.91} & \textbf{6.00} & \textbf{6.11} & \textbf{6.21} \\
		\hline
	\end{tabular}
\end{table}

\textbf{Evaluation on Model Features:}
Besides the ability of generalizing the faces, DAMs can produce a higher level representation for both facial shape and texture. Therefore, instead of using pixel values, we extracted the model parameters as described in Section \ref{subsec:DAMProp} and evaluated them with the age estimation system. For the AAMs features, the number of features for shape and texture was chosen so that $93\%$ of variations are retained.
Table \ref{table_MAE_age_estimation_comparison} lists the MAEs of four different inputs: reconstructed image of DAMs (\textbf{DAMs-Rec}) and AAMs (\textbf{AAMs-Rec}), model parameters extracted from AAMs (\textbf{AAMs-Mod}) and DAMs (\textbf{DAMs-Mod}) as well as other age estimation methods.
Not surprisingly, our DAMs feature achieves the lowest MAEs as compared with AAMs features.
Notice that, in this experiment, although DAMs are not tuned toward the aging labels, the features extracted by DAMs still achieve quite competitive results to others. We believe that with better age estimator, i.e. deep learning based age estimator, and the use of aging labels during training DAMs, the MAE would be reduced significantly. We leave this as our future work of DAMs.

\begin{table}[!t]
	\caption{Comparison of age estimation results on FG-NET database with four different features and other age estimation approaches.}
	\label{table_MAE_age_estimation_comparison}
	\centering
	\begin{tabular}{|c|c|}
		\hline
		\textbf{Inputs} & \textbf{MAEs (years)} \\
		\hline
		\hline
		DAMs-Mod & \textbf{4.67} \\
		\hline
		AAMs-Mod & 4.81 \\
		\hline
		DAMs-Rec & 5.67 \\
		\hline
		AAMs-Rec & 6.14 \\
		\hline
		\hline
		DLF-CNN ~\citep{wang2015deeply} & 4.26 \\
		\hline
		CA-SVR ~\citep{chen2013cumulative} & 4.67 \\
		\hline
		PLO ~\citep{li2012learning} & 4.82 \\
		\hline
		
	\end{tabular}
\end{table}

\subsection{Shape Fitting in DAMs} \label{subsec:Shape_fitting}
\begin{figure}[t]
	\begin{center}
		\includegraphics[width=6.5cm]{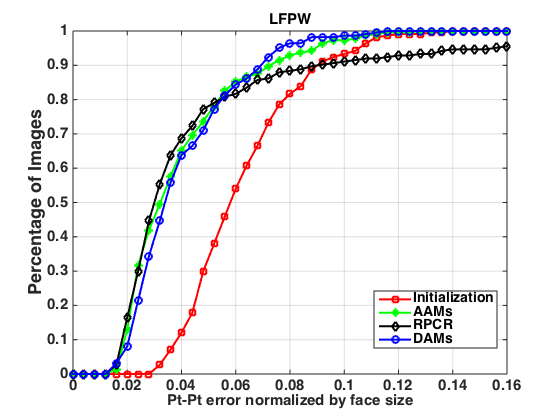}
	\end{center}
	\caption{Cumulative Error Distribution (CED) curves of LFPW database.}
	\label{fig:CS_score_lfpw}
\end{figure}

Besides some other previous shape fitting approaches (\cite{alabort2014menpo,alabort2014bayesian,alabort2015unifying,alabort2017unified,antonakos2015feature, tzimiropoulos2017fast}), we employ our proposed DAMs to this problem on LFPW database to further demonstrate their robustness and effectiveness.
The model configurations are kept the same as in previous sections except it is now trained with 811 training images of LFPW.
For evaluation and comparison, we use the average distance of each landmark to its ground truth position normalized by face size as in ~\citep{tzimiropoulos2013optimization}.
Moreover, in order to remove the effect of face detection error during fitting step, we use the bounding boxes provided in ~\citep{sagonas2013semi} for initialization. Then we simply place the mean shape with 68 landmarks inside the face's bounding box and start the fitting process.
We compare our method with two other fitting strategies, i.e. AAMs and RCPR ~\citep{burgos2013robust}, and present the results in Table \ref{table_DAMs_fitting}. The Cumulative Error Distribution (CED) curves are showed in Figure \ref{fig:CS_score_lfpw}.

\textbf{Comparisons Against AAMs Based Approaches}: We conduct an ablation study to compare between our DAMs and other AAMs based fitting approaches in Table \ref{table_DAMs_fitting_compare_feature}. In these comparisons, we also include the feature-based AAMs ~\citep{antonakos2015feature}, i.e. represent texture with HoG and Dense SIFT features, as well as hybrid approaches, i.e. Supervised Descent Method combined with AAMs (SDM + AAMs), proposed in ~\citep{antonakos2016adaptive}. Notice that to make a fair comparison between our DAMs and other approaches, we use a single-resolution pyramidal scheme for the AAMs model configurations. The Project out Forward Compositional (POFC) fitting algorithm is used in this experiment. For holistic AAMs, the number of appearance parameters is fixed at 50. The dimension of 2D shape parameters is set to 12. The results again show that our DAMs achieve a comparable fitting error with feature-base AAMs.

\begin{table}[t]
	\caption{The fitting errors using different methods against LFPW database.}
	\label{table_DAMs_fitting}
	\centering
	\begin{tabular}{|c|c|}
		\hline
		\textbf{Methods} & \textbf{Fitting Error} \\
		\hline
		Initialization & 0.0618 \\
		\hline
		Fast-SIC 
		& 0.0391\\
		\hline
		RCPR 
		& 0.0505\\
		\hline
		\hline
		DAMs (Ours) & 0.0398\\
		\hline
	\end{tabular}
\end{table}

\begin{table}[t]
	\caption{The fitting errors against other feature-based AAMs methods and hybrid approach (SDM + AAMs).}
	\label{table_DAMs_fitting_compare_feature}
	\centering
	\begin{tabular}{|c|c|}
		\hline
		\textbf{Methods} & \textbf{Fitting Error} \\
		\hline
		HoG AAMs 
		& 0.0372\\
		\hline
		Dense SIFT AAMs
		& 0.035\\
		\hline
        \hline
		SDM + AAMs & 0.0407 \\
        \hline
		\hline
		DAMs (Ours) & 0.0398\\
		\hline
	\end{tabular}
\end{table}

\begin{figure}[t]
	\begin{center}
		\includegraphics[width=6cm]{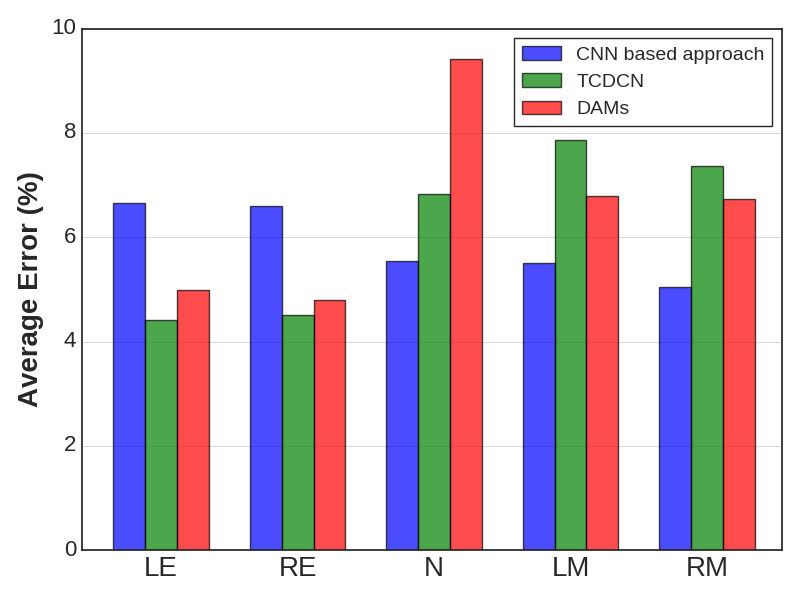}
	\end{center}
	\caption{The average errors (\%) of CNN based approach ~\citep{sun2013deep}, TCDCN ~\citep{zhang2016learning}, and DAMs of different landmarks: left eye (LE), right eye (RE), nose (N), left and right mouth corners (LM and RM).}
	\label{fig:average_error_LMK_detect}
\end{figure}

\textbf{Comparisons Against CNN Based Approaches}: We also compare our method with another CNN based landmark detection approach ~\citep{sun2013deep}, and Tasks-Constrained Deep Convolutional Network (TCDCN) ~\citep{zhang2016learning}. Since these approaches results in a detection of five landmark points (i.e. left eye, right eye, nose, and two mouth corners), we only compute the average distance of these landmarks for comparison. Our DAMs approach achieves the fitting error of 0.028 while the error of the CNN based system and TCDCN are 0.026 and 0.027, respectively. The average detection errors ~\citep{sun2013deep} of all five landmarks are also presented in Figure \ref{fig:average_error_LMK_detect}.
These results show that DAMs achieve comparable accuracy to other face alignment methods.

\textbf{The Sensitivity Against Different Shape Initial Bounding Boxes}: In order to evaluate the sensitivity of our DAMs approach to the size of detected bounding boxes during fitting step, we employ various face detection techniques including the initial bounding boxes provided by ~\citep{sagonas2013semi}, MTCNN ~\citep{zhang2016joint}, CMS-RCNN ~\citep{zhu2017cms} to obtain different types of bounding boxes. Then, we place the mean shape inside these bounding boxes and start the fitting process. Table \ref{table_DAMs_fitting_initialization} presents the fitting accuracy of our DAMs fitting approach when different bounding boxes are used. From these results, one can see that our fitting approach is quite robust to the initial location and scale of the mean shape.

\begin{table}[t]
	\caption{The fitting errors against different initial bounding boxes.}
	\label{table_DAMs_fitting_initialization}
	\centering
	\begin{tabular}{|c|c|}
		\hline
		\textbf{Face Detection Methods} & \textbf{DAMs Fitting Error} \\
		\hline
		Initialization from ~\citep{sagonas2013semi} & 0.0398 \\
		\hline
		MTCNN ~\citep{zhang2016joint}
		& 0.0383\\
		\hline
		CMS-RCNN ~\citep{zhu2017cms}
		& 0.0395\\
		\hline
	\end{tabular}
\end{table}

\begin{table}[t]
	\caption{Computational time of DAMs and AAMs in three stages: training the models, fitting and reconstruct an image.}
	\label{table_computational_cost}
	\centering
	\begin{tabular}{|l|c|c|c|}
		\hline
		\textbf{Stages} & \textbf{Training} & \textbf{Fitting} & \textbf{Reconstruction}\\
		\hline
		\hline
		DAMs & 12.87 hrs & 17.5 s & 0.53 s \\
		\hline
		AAMs 
		& 564.06 s & 2.28 s & 0.023 s\\
		\hline
	\end{tabular}
\end{table}

\subsection{Computational Costs} \label{subsec:Computationtal_cost}
The computational costs of DAMs, i.e. training, fitting and reconstruction stages are discussed in this section. Both Helen and LFPW databases are combined to use in this evaluation.
The numbers of training and testing images are 2811 and 554, respectively. The method is implemented in Matlab environment and runs in a system of Core i7-2600 @3.4GHz CPU, 8.00 GB RAM.
The shape contains 68 landmarks and the appearance is represented in a vector of 9652 dimensions. Each layer was trained using Contrastive Divergence learning in 600 epochs.
It is noted that the current version is implemented without using parallel processing. The computational costs of DAMs and AAMs are shown in Table \ref{table_computational_cost}.

\section{Conclusions} \label{sec:Conclusion}
This paper has introduced novel Deep Appearance Models that have abilities of generalizing and representing faces in large variations.
With the deep structured models for shapes and textures, the proposed approach was shown to achieve remarkable improvements in both facial reconstruction and facial age estimation tasks compared with PCA-based AAMs model.
Moreover, the new model can produce a more robust face shape and texture representation based on their high-level relationships.
Experimental results in several applications such as facial super-resolution, face off-angle reconstruction, occlusion removal and facial age estimation have shown the potential of the model in dealing with large variations.

\begin{acknowledgements}
This work was supported in part by the Natural Sciences
and Engineering Research Council (NSERC) of Canada.
\end{acknowledgements}

\bibliographystyle{spbasic}      
\bibliography{egbib.bib}   

\end{document}